\documentclass{article}

\PassOptionsToPackage{numbers,compress}{natbib}

 \usepackage[preprint]{neurips_2026}

\usepackage[utf8]{inputenc} 
\usepackage[T1]{fontenc}    
\usepackage{hyperref}       
\usepackage{url}            
\usepackage{booktabs}       
\usepackage{amsfonts}       
\usepackage{nicefrac}       
\usepackage{microtype}      
\usepackage{xcolor}         
\usepackage{amssymb}
\usepackage{graphicx}
\usepackage{float}
\usepackage{subcaption}
\usepackage{booktabs} 
\usepackage{multirow}
\usepackage{array}
\newcolumntype{P}[1]{>{\raggedright\arraybackslash}p{#1}}
\usepackage{graphicx}
\usepackage{pifont}
\usepackage{caption}
\usepackage{wrapfig}
\usepackage{enumitem}
\usepackage[table]{xcolor}
\usepackage{amsmath}
\usepackage{algorithm}
\usepackage{algorithmic}
\usepackage{makecell}
\newcommand{\fixgood}[1]{\textit{#1}}
\newcommand{\fixbad}[1]{\textcolor{red}{\underline{#1}}}
\newcommand{\stillwrong}[1]{\textcolor{blue}{\underline{#1}}}

\title{What Are We Actually Decoding? Source Attribution for Non-Invasive Brain-to-Language Retrieval}

\author{
\textbf{Xinyu Zhang\textsuperscript{1,3,4$\dagger$}} \quad
\textbf{Sichao Liu\textsuperscript{1,2,3$\dagger$,$\ddagger$ }} \quad
\textbf{Runhao Lu\textsuperscript{2,5}} \quad
\textbf{Alexandra Woolgar\textsuperscript{2}} \quad 
\textbf{Lihui Wang\textsuperscript{1}} \\
\textsuperscript{1}KTH, Sweden \quad
\textsuperscript{2}University of Cambridge, UK \quad
\textsuperscript{3}EPFL, Switzerland \quad \\
\textsuperscript{4}Karolinska Institutet, Sweden \quad
\textsuperscript{5}McGill University, Canada
}

\begin{document}

\maketitle
\begingroup
\renewcommand\thefootnote{}
\footnotetext{$\dagger$: equal contributors; $\ddagger$ Corresponding author: \url{sicliu@kth.se}}
\endgroup

\begin{abstract}
In non-invasive neural language decoding, results can be inflated by sources that are not stimulus-evoked neural evidence: decoder priors, embedding-based metrics, and non-neural structural nuisances such as signal duration. The methodological challenge is therefore attribution: a reported gain is more informative when it can be traced to a specific source. We recast stimulus-locked MEG-to-audio retrieval as an auditing framework that separates apparent performance into three sources --- structural shortcuts, window-level stimulus-locked evidence, and cross-window contextual aggregation --- and provides a diagnostic for each. Signal-blind Gaussian noise reaches 66.3\% Rank@1 (R@1) under variable-length decoding but collapses to near chance once fixed-duration windows and stimulus-identity splits are enforced, isolating structural leakage. Under these controls, fixed-window retrieval recovers measurable MEG--audio discriminability, while an oracle sentence-bucket diagnostic shows that 95.7\% of Top-1 errors select the wrong sentence, localising the residual bottleneck to sentence-level competition. We audit this contextual source with Group Context Bias (GCB), an inference-time additive logit bias that pools sentence-consistent evidence across windows while leaving the base retrieval scores and candidate pool fixed. Used as a score-space intervention, GCB makes the contextual source measurable: R@1 shifts from 44\% to 52\% on Gwilliams and from 22\% to 29\% on MOUS under the same fixed setting. GCB is auditable under this design: its effect collapses under random-grouping perturbations and vanishes when local evidence is attenuated in MEG or is near chance in EEG, supporting its use as a controlled source-attribution intervention. These results suggest that brain-to-language performance should be source-attributed, not merely reported. 
\end{abstract}

\section{Introduction}
\label{sec:intro}
Non-invasive neural language decoding has advanced rapidly with the introduction of pretrained speech and language models, and end-to-end systems increasingly report strong performance on text generation and semantic matching benchmarks~\citep{wang2022open,duan2023dewave,xi2023unicorn,yang2024mad,li2025brainecho}.
Yet reported accuracy in this setting is often not what it appears to be. Under a standard variable-length sentence-decoding protocol, \textbf{replacing real MEG with Gaussian noise still yields 66.3\% next-token R@1 accuracy} --- variable-duration inputs introduce a sequence-length shortcut~\citep{geirhos2020shortcut, baillargeon2022assessing} that attention-based models exploit through padding boundaries and mask patterns. Duration is one instance of a broader pattern: \textbf{decoder priors}~\citep{jo2026evaluating}, \textbf{embedding-based metrics}~\citep{zhang2019bertscore,hanna2021fine}, \textbf{training-evaluation leakage}~\citep{brookshire2024data,yin2025rethinking}, and other \textbf{non-neural structural nuisances} can each make signal-blind inputs appear informative, so reported gains can reflect evaluation shortcuts rather than decodable linguistic information in the neural signal. The \textbf{methodological challenge} is therefore attribution: \textbf{how can we verify whether apparent brain-to-language decoding performance is genuinely supported by stimulus-locked neural evidence, rather than by structural shortcuts, decoder priors, or evaluation artefacts?}
 
Text generation is a poor endpoint for this kind of attribution. Neural evidence, decoder priors, decoding strategy, and text-level metrics are all entangled in the final output, so a high semantic score or fluent generation alone does not identify the source of the information being used. This ambiguity is especially acute when a neural encoder conditions a strong speech, language, or generative prior: plausible outputs can be produced even when the neural representation contributes little stimulus-specific evidence. Stimulus-locked MEG-to-audio retrieval offers a more controlled endpoint~\citep{defossez2023decoding}: each neural window retrieves its time-aligned audio window from an explicit candidate pool, making the candidate set, target alignment, chance level, and rank errors directly observable. Retrieval therefore turns apparent decoding performance into auditable candidate-level decisions, allowing structural shortcuts, local stimulus-locked evidence, and contextual effects to be analysed separately.

Within this auditable setting, retrieval performance arises from multiple sources, and we trace them in turn. To isolate window-level stimulus-locked evidence, we first use independent fixed-window retrieval: each window is decoded separately using fixed-duration windows and stimulus-identity splits, preventing variable duration and repeated stimulus content from acting as direct ranking cues~\citep{brookshire2024data,yin2025rethinking,defossez2023decoding,d2025towards}. This local baseline shows MEG--audio discriminability, indicating that the window-level stimulus-locked neural signal is decodable under the controlled retrieval setting.

However, this local-only design factorises retrieval over individual windows and therefore does not condition on information beyond the current window, despite evidence that linguistic context modulates time-locked neural responses during natural speech~\citep{slaats2023delta,brodbeck2022parallel,raghavan2025neural}. The error structure confirms this gap: under a global candidate pool, 95.7\% of Top-1 errors select a candidate from the wrong sentence, and an oracle diagnostic that restricts candidates to the ground-truth sentence raises R@1 from 44\% to 88\% with the model held fixed. Many failures, therefore, reflect unresolved sentence-bucket competition, where useful cross-window structure in the local logits is left unused by independent decoding. Cross-window contextual aggregation, therefore, emerges as a third performance source — one that should be modelled and attributed \emph{separately} from local evidence rather than absorbed into a stronger encoder. We study it through an explicit logit-space correction, keeping the contextual contribution structurally decoupled from the frozen retrieval model.

Together, these contributions form an attribution methodology for non-invasive neural language decoding. Concretely:
\begin{itemize}[itemsep=0pt, topsep=0pt, leftmargin=*]
\item \textbf{Quantifying structural nuisances.} 
We show that variable-duration decoding enables a sequence-length shortcut: in a sentence-level setup, signal-blind Gaussian noise yields 66.3\% R@1 because duration, padding, or mask structure remains informative without neural content. Fixed-duration retrieval windows and stimulus-identity splits collapse this shortcut and repeated-stimulus leakage to near chance, providing a retrieval endpoint that is largely independent of non-neural structure.

\item \textbf{Isolating window-level stimulus-locked evidence.}
Under these controls, independent fixed-window retrieval recovers measurable MEG--audio discriminability across two datasets, isolating window-level evidence as a quantifiable source of decoding performance before any contextual aggregation is applied.

\item \textbf{Auditing cross-window contextual aggregation.}
Beyond the oracle diagnostic, we audit the contextual source using GCB, a logit-space correction that leaves the frozen retrieval model, candidate pool, and within-bucket ordering untouched. Under frozen local logits, this intervention estimates the recoverable contextual contribution: R@1 shifts from 44\% to 52\% on Gwilliams and from 22\% to 29\% on MOUS under the same fixed setting, and the contrast vanishes under random grouping or evidence-limited controls.

\end{itemize}

\begin{figure*}[!ht]
  \centering
  \includegraphics[width=1.0\linewidth]{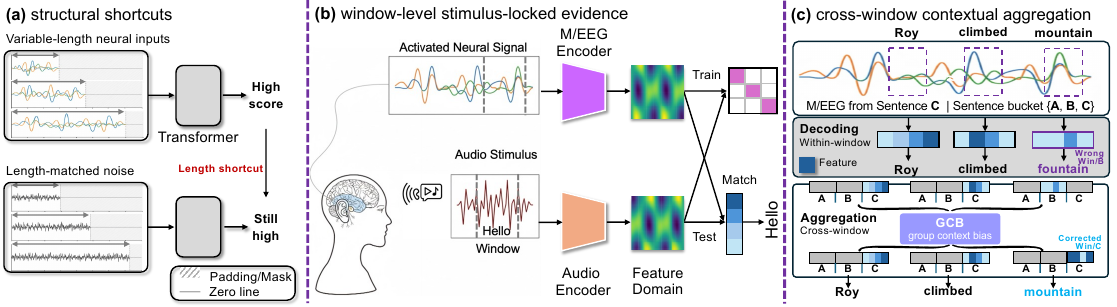}
    \caption{\textbf{Three performance sources in stimulus-locked M/EEG-to-audio retrieval.}
    (a) \textbf{Representative structural shortcut.}
    Variable-duration inputs provide one concrete example of a structural shortcut: when neural activity is replaced by duration-preserving Gaussian noise, above-chance decoding indicates reliance on signal length or padding structure rather than neural content.
    (b) Fixed-duration, stimulus-locked M/EEG windows and aligned audio windows are encoded into a shared feature space with a contrastive objective, yielding an independent window-level retrieval baseline.
    (c) Cross-window contextual aggregation is isolated as a separate score-space intervention. Each window first produces local retrieval logits over the full candidate pool, organised here into sentence buckets \(\{A,B,C\}\), and independent decoding can yield wrong-sentence near misses, such as a sentence-\(C\) window selecting a bucket-\(B\) candidate. GCB pools the fixed local logits across windows from the same query sentence to estimate bucket-level support, then adds a bucket-wise bias to the original logits. In this example, aggregated evidence supports bucket \(C\), correcting the cross-sentence error while leaving the encoders, candidate pool, and within-bucket ordering unchanged.}
  \vspace{-1.2em}
  \label{fig:gcb_overview}
\end{figure*}

\section{Related Work}
\label{sec:related}
\textbf{End-to-end brain-to-text decoding and evaluation validity.}
Recent non-invasive brain-to-text systems combine neural encoders with pretrained speech or language models through direct generation~\citep{wang2022open,xi2023unicorn,yang2024mad}, discrete intermediate representations~\citep{duan2023dewave,li2025brainecho}, and semantic-alignment objectives~\citep{feng2023aligning,zhou2024belt,liu2024eeg2text}.
Yet several lines of analysis show that reported gains in this setting can arise from sources other than neural evidence.
Noise-input and teacher-forcing controls reveal that decoder priors and protocol choices inflate apparent decoding performance~\citep{jo2026evaluating}.
Inappropriate train--test splitting overestimates performance in EEG deep learning and brain-to-text~\citep{brookshire2024data,yin2025rethinking}.
Embedding-based metrics such as BERTScore reward semantic similarity rather than exact stimulus recovery, obscuring attribution under low SNR~\citep{zhang2019bertscore,hanna2021fine}.
Each concern reflects the broader pattern that benchmarks are routinely inflated by shortcut learning and annotation artefacts~\citep{geirhos2020shortcut,mccoy2019right,gururangan2018annotation}, and together they motivate a shift from comparing final scores to auditing where they come from.

\textbf{Retrieval-based decoding and controlled evaluation.}
Closed-set retrieval addresses these validity concerns by making chance performance explicit and failure modes observable.
\citet{defossez2023decoding} established a strong M/EEG-to-audio retrieval formulation by contrastively aligning neural windows with pretrained speech representations, drawing on CLIP-style contrastive learning~\citep{radford2021learning} and InfoNCE-based representation learning~\citep{chen2020simple}.
Subsequent work extends retrieval to word-level decoding and stronger architectures~\citep{d2025towards,boyko2024megformer,pmlr-v267-jayalath25a,zhou2024towards}, and to other stimulus modalities~\citep{benchetrit2023brain,song2023decoding,rajabi2025human}.
We adopt the exp-dilated CNN of \citet{defossez2023decoding} as a controlled reference, and turn retrieval into an \emph{auditable} attribution endpoint through fixed-duration windows, stimulus-identity splits, and rank-error diagnostics.

\textbf{Context modelling and score-space aggregation.}
Electrophysiological studies show that linguistic context modulates time-locked neural responses during natural speech~\citep{slaats2023delta,brodbeck2022parallel,raghavan2025neural,broderick2018semantic}.
Recent decoding work exploits this through training-time mechanisms, such as sentence-level Transformers over local CNN features~\citep{d2025towards}.
These approaches can improve decoding, but they entangle local representation learning with contextual integration, making source attribution less direct.
In information retrieval, by contrast, rank fusion and score composition combine evidence at inference time without retraining~\citep{cormack2009reciprocal,cserbetcci2024hu,faltings2025enhancing}.
GCB brings this inference-time principle to neural retrieval: it pools sentence-consistent evidence across neighbouring windows in logit space, keeping contextual correction structurally separable from the frozen retrieval model so that local and contextual contributions can be audited independently.

\section{Methods}
\label{sec:methods}

\subsection{Stimulus-locked retrieval as an attribution endpoint}
\label{subsec:mtd_task}

We evaluate non-invasive neural language decoding via fixed-duration, stimulus-locked M/EEG-to-audio retrieval. Each query is a neural window aligned to a presented speech segment; each candidate is an audio window from an explicit inference-time pool.

Let \(\mathcal{Q}\) be the set of query neural windows and \(\mathcal{C}\) the candidate pool, whose size \(N=|\mathcal{C}|\) depends on the evaluation setting; for each \(w\in\mathcal{Q}\), let \(j^\star(w)\in\mathcal{C}\) be the time-aligned target. Each window \(u\) carries a stimulus-identity key \(\kappa(u)\) (used in Sec.~\ref{subsec:mtd_controls} for split control) and a sentence identity \(\sigma(u)\) (used by the diagnostic of Sec.~\ref{subsec:mtd_controls} and the aggregation of Sec.~\ref{subsec:mtd_context}). Candidates are de-duplicated by stimulus identity and window index, so each candidate is a unique audio segment and chance is well-defined. Sentence identities are not part of the local retrieval definition: logits \(\ell_{w,j}\) are computed for all pairs in \(\mathcal{Q}\times\mathcal{C}\), and sentence buckets enter only post hoc.

The retrieval rank \(r(w)\) is the position of \(j^\star(w)\) after sorting \(\{\ell_{w,j}\}_{j\in\mathcal{C}}\) in descending order. We report Rank@\(K\) (R@K), mean reciprocal rank (MRR), and median rank (MedR):

\begin{equation}
\mathrm{R@}K = \tfrac{1}{|\mathcal{Q}|}\!\sum_{w\in\mathcal{Q}}\!\mathbf{1}[r(w)\le K],\quad
\mathrm{MRR} = \tfrac{1}{|\mathcal{Q}|}\!\sum_{w\in\mathcal{Q}}\!\tfrac{1}{r(w)},\quad
\mathrm{MedR} = \mathrm{median}\{r(w):w\in\mathcal{Q}\}.
\label{eq:retrieval_metrics}
\end{equation}

\subsection{Controlling structural shortcuts and localising contextual headroom}
\label{subsec:mtd_controls}

We first control variable signal duration and repeated stimulus content at the retrieval endpoint, then use an oracle diagnostic to localise the residual contextual headroom.

\textbf{Fixed-duration stimulus-locked windows.}
For a word onset \(t_w\), the neural input is
\begin{equation}
x_w = X[t_w+\delta-0.5,\;t_w+\delta+2.5]\in\mathbb{R}^{C\times T},
\label{eq:window_def}
\end{equation}
with neural lag \(\delta=150\) ms (matched to typical M/EEG auditory response latency), \(C\) sensors, and \(T=360\) at 120\,Hz~\citep{defossez2023decoding}. This follows the standard event-related paradigm in M/EEG language research~\citep{luck2014introduction,beres2017time,gwilliams2023introducing} and removes the duration cue at the input level, rather than encoding it implicitly through padding or attention masks.

\textbf{Stimulus-identity splits.}
The same stimulus content can appear across subjects, sessions, and repetitions; a random split can therefore place different observations of the same content in both the training and test sets.
We therefore split by stimulus identity, keyed on what was heard rather than who heard it.
Concretely, all windows sharing the same stimulus-identity key are assigned to the same training, validation, or test split.
This prevents repeated observations of the same stimulus window from crossing partitions.

\textbf{Oracle sentence-bucket diagnostic.}
With duration and identity confounds controlled, we localise the dominant remaining error source. For each query \(w\), let \(\mathcal{B}_{s^\star} = \{j\in\mathcal{C}:\sigma(j)=\sigma(j^\star(w))\}\) be the candidate bucket containing the target. Ranking \(j^\star(w)\) using the same logits restricted to \(\mathcal{B}_{s^\star}\) --- with the model, logits, and target candidate held fixed --- separates within-bucket discriminability from cross-bucket competition. The gap between full-pool and within-bucket rank identifies contextual headroom beyond isolated window-level retrieval, motivating the intervention of Sec.~\ref{subsec:mtd_context}.

\subsection{Local neural-to-audio retrieval model}
\label{subsec:mtd_local}

\textbf{Retrieval scoring.}
Each audio candidate is encoded by a frozen speech encoder~\citep{baevski2020wav2vec} into a time-resolved target
\(y^{\mathrm{aud}}_j\in\mathbb{R}^{D\times T}\), shared across datasets
(Appendix~\ref{app:audio_embed_details}).
A neural encoder \(f_\theta\) maps each neural window \(x_w\) to
\(z^{\mathrm{neu}}_w=f_\theta(x_w)\in\mathbb{R}^{D\times T}\).
Retrieval logits are Frobenius inner products over the \(D\times T\) matrices, i.e., summed element-wise products over feature and time.
We denote this inner product by \(\langle\cdot,\cdot\rangle_F\), and normalise audio targets by the Frobenius norm \(\|\cdot\|_F\) using a small \(\epsilon>0\) for numerical stability:
\begin{equation}
\tilde{y}^{\mathrm{aud}}_j
=
\frac{y^{\mathrm{aud}}_j}{\|y^{\mathrm{aud}}_j\|_F+\epsilon},
\qquad
\ell_{w,j}
=
\left\langle z^{\mathrm{neu}}_w,\tilde{y}^{\mathrm{aud}}_j\right\rangle_F .
\label{eq:retrieval_logits}
\end{equation}

The \(D\times T\) scoring preserves time-resolved alignment across the 3\,s window.
We optimise \(f_\theta\) with a neural-to-audio contrastive retrieval loss: in each batch of size \(B\), the matched logit \(\ell_{i,i}\) competes against all audio candidates \(\{\ell_{i,j}\}_{j=1}^{B}\):
\begin{equation}
\mathcal{L}
=
-\frac{1}{B}
\sum_{i=1}^{B}
\log
\frac{\exp(\ell_{i,i})}
{\sum_{j=1}^{B}\exp(\ell_{i,j})}.
\label{eq:loss}
\end{equation}

\textbf{Encoder instantiations.}
The neural encoder follows a shared retrieval pipeline: coordinate-conditioned spatial mixing maps each native M/EEG sensor layout to a common latent spatial grid, a subject-specific layer adapts the shared representation across participants, and a temporal convolutional backbone maps each window to the audio-target space.
Within this fixed pipeline, we compare three temporal-backbone instantiations: (i) Dense-TCNN, (ii) a receptive-field-matched exp-dilated CNN and (iii) an exp-dilated variant adapted from~\citet{defossez2023decoding}.
All instantiations use the same preprocessing, windowing, candidate construction, audio targets, retrieval loss, optimiser, and model-selection protocol; architectural details and receptive-field calculations are given in Appendix~\ref{app:arch}.
The resulting local logit matrix \(\ell_{w,j}\) is treated as fixed local evidence; Sec.~\ref{subsec:mtd_context} operates on top of it without modifying the encoder, audio embeddings, or candidate pool.

\subsection{Contextual aggregation as a separable score-space intervention}
\label{subsec:mtd_context}

GCB instantiates the third source as an inference-time score-space intervention on the fixed local logits from Sec.~\ref{subsec:mtd_local}.
The local retrieval model decodes each window independently, even when neighbouring windows from the same sentence carry consistent evidence.
GCB adds cross-window contextual information only through an additive bucket-wise bias, while keeping the neural encoder, audio embeddings, candidate pool, and base logits fixed.
Thus, the Base--GCB contrast isolates the effect of this score correction.

\textbf{GCB definition.}
Let \(L\in\mathbb{R}^{B\times N}\) be the local-logit matrix for \(B\) query windows and \(N\) candidate audio windows.
Sentence identity partitions the candidate side into \(M\) buckets.
Let \(c(j)\in\{1,\dots,M\}\) denote the bucket index of candidate \(j\); the candidate set in bucket \(s\) is
\begin{equation}
\mathcal{B}_s = \{j\in[1,N]:c(j)=s\}.
\label{eq:gcb_bucket}
\end{equation}
Query windows are grouped by the presented sentence; let \(G\subseteq\{1,\dots,B\}\) denote one such query group.
GCB estimates which candidate buckets receive consistent support across \(G\) and adds a bucket-wise bias to candidates in the most supported buckets.

\textbf{Local evidence extraction.}
For each \(b\in G\), GCB retains high-scoring local candidates. Let \(\mathcal{I}_b=\mathrm{TopK}(L_{b,:},K)\) and \(\tau_b=\mathrm{Quantile}_q(L_{b,:})\) over the full candidate set; the retained set and excess scores are
\begin{equation}
\mathcal{J}_b = \{j\in\mathcal{I}_b:L_{b,j}\ge\tau_b\},\qquad
e_{b,j} = L_{b,j}-\tau_b.
\label{eq:gcb_evidence}
\end{equation}

\textbf{Bucket support.}
Evidence is pooled across \(G\) at the bucket level.
Here, \(\operatorname{MeanTop}_{m}\) averages the largest \(m\) retained scores in a set, \(\operatorname{Top}_{S}\) returns the indices of the \(S\) largest bucket-support values, and \(\mathbf{1}[\cdot]\) denotes an indicator.
For bucket \(s\), let
\begin{equation}
\mathcal{E}_G(s)
=
\{e_{b,j}: b\in G,\; j\in\mathcal{J}_b,\; c(j)=s\}
\label{eq:gcb_Edef}
\end{equation}
be the retained evidence assigned to that bucket.
Its support is
\begin{equation}
U_G(s)
=
\mathrm{norm}(|\mathcal{B}_s|)
\cdot
\operatorname{MeanTop}_{m}\!\left(\mathcal{E}_G(s)\right),
\label{eq:gcb_support}
\end{equation}
with \(U_G(s)=0\) when \(\mathcal{E}_G(s)=\emptyset\).
The normaliser \(\mathrm{norm}(|\mathcal{B}_s|)\) controls bucket-size exposure, preventing larger sentence buckets from gaining support merely because they contain more candidates.

\textbf{Score correction.}
Let
\(\mathcal{S}_G=\operatorname{Top}_{S}(\{U_G(s)\}_{s=1}^{M})\)
be the selected high-support buckets.
For \(b\in G\) and \(j\in[1,N]\),
\begin{equation}
L^{\mathrm{GCB}}_{b,j}
=
L_{b,j}
+
\gamma_{\mathrm{gcb}}\,U_G(c(j))\,
\mathbf{1}\!\left[c(j)\in\mathcal{S}_G\right].
\label{eq:gcb_bias}
\end{equation}
The scale \(\gamma_{\mathrm{gcb}}\) sets the bias strength. The correction is bucket-constant, and GCB preserves within-bucket ordering and changes only cross-bucket competition. Hyperparameters \((K,q,m,S,\gamma_{\mathrm{gcb}},\mathrm{norm})\) are selected on Gwilliams validation split and fixed for all test evaluations.

\textbf{Attribution controls.}
Complementary controls test which part of the intervention produces the observed Base--GCB contrast:
GCB-single removes cross-window pooling; hard pruning tests candidate-pool restriction; grouping perturbations test sentence-structure dependence; and evidence-limited controls (within-Gwilliams attenuation; Brennan EEG) test whether the contextual effect remains observable when local stimulus-locked evidence is weakened.
A training-time local--global coupling variant is reported (Appendix~\ref{app:coupling}).

\section{Experiments}
\label{sec:exp}

\vspace{-0.5em}
\subsection{Experimental setup}
\label{subsec:exp_setup}
\vspace{-0.5em}

\textbf{Datasets.}
We evaluate three naturalistic speech-perception datasets (Table~\ref{tab:dataset_overview_setup}), each serving a distinct role in the attribution analysis.
Gwilliams is the primary setting for isolating structural shortcuts, window-level stimulus-locked evidence, and cross-window contextual aggregation under shortcut-controlled retrieval.
MOUS tests whether the same attribution procedure transfers across language and recording conditions without retuning.
Brennan EEG serves as an evidence-limited negative control, complemented by a within-Gwilliams attenuation diagnostic that weakens local stimulus-locked evidence while holding the dataset, candidate pool, grouping structure, encoder, and GCB hyperparameters fixed (Sec.~\ref{subsec:res_additional}).
We report closed-set retrieval in a zero-shot held-out setting (candidate size: \( N_{\mathrm{zs}}\)) and, for Gwilliams only, a session-isolated setting with earlier-session training and later-session evaluation over a larger pool (\(N_{\mathrm{si}}\); Appendix~\ref{app:slr}).

\begin{wraptable}{r}{0.50\textwidth}
\vspace{-1em}
\centering
\small
\setlength{\tabcolsep}{2pt}
\caption{\textbf{Dataset overview and closed-pool evaluation sizes.} 
\(N_{\text{zs}}\): zero-shot; \(N_{\text{si}}\): session isolation.}
\label{tab:dataset_overview_setup}
\resizebox{0.5\textwidth}{!}{
\begin{tabular}{lcccccc}
\toprule
Dataset & Language & Mode & Sensors & Subjects & Duration & \(N_{\text{zs}}/N_{\text{si}}\) \\
\midrule
\textbf{Brennan}   & EN & EEG & 60  & 33 & 6.7\,h  & 388/-- \\
\textbf{MOUS}      & NL & MEG & 272 & 96 & 80.9\,h & 825/-- \\
\textbf{Gwilliams} & EN & MEG & 208 & 27 & 56.2\,h & 1{,}464/7{,}011 \\
\bottomrule
\end{tabular}
}
\vspace{-1em}
\end{wraptable}

\textbf{Preprocessing.}
All compared methods use the same stimulus-side timing to define sentence boundaries and word onsets.
Preprocessing is shortcut-controlled and shared across datasets and encoder variants; in particular, scaling statistics are computed on the training split only.
Dataset-specific window construction and preprocessing details are given in Appendices~\ref{app:meta_manifest}, \ref{app:mous_whisper}, and~\ref{app:preproc}.

\textbf{Training and evaluation.}
Base encoders are trained following Sec.~\ref{subsec:mtd_local}, with checkpoints selected by validation loss.
Zero-shot experiments evaluate stimulus-identity-disjoint test windows over three random seeds; Gwilliams session-isolation evaluates held-out recording sessions over five folds \(\times\) three seeds.
GCB hyperparameters are selected on the Gwilliams validation split and fixed for all test evaluations, including MOUS transfer.
Unless stated otherwise, the fixed GCB setting is \(K{=}128\), \(q{=}0.95\), \(m{=}3\), \(S{=}3\), bucket-sqrt normalisation, and \(\gamma_{\mathrm{gcb}}{=}0.7\).
Full architecture and training details are reported in Appendix~\ref{app:arch}; audio-embedding and GCB-sweep details are reported in Appendices~\ref{app:audio_embed_details} and~\ref{app:gcb_scan}.

\textbf{Attribution contrasts.}
For each temporal backbone in Sec.~\ref{subsec:mtd_local}, we compare the frozen local retrieval logits before and after applying GCB, so the Base--GCB contrast isolates the effect of the inference-time score correction.
Component, perturbation, evidence-limited, and architectural controls are reported in Sec.~\ref{subsec:res_additional} and Appendix~\ref{app:mechanism}.
A Whisper-based next-token prediction (Appendix~\ref{app:gen_details}) is used to expose the variable-duration shortcut.

\vspace{-1em}
\subsection{Experimental results and analysis}
\label{sec:experiments}
\vspace{-0.5em}

We trace the three performance sources: (i) structural shortcuts, (ii) window-level stimulus-locked evidence under retrieval, and (iii) cross-window contextual aggregation through GCB.

\textbf{Source 1: structural shortcuts and text-side metrics inflate apparent decoding.}
A Whisper-based next-token diagnostic on Gwilliams (Appendix~\ref{app:gen_details}) exposes a duration shortcut, and a separate embedding-metric diagnostic shows that text similarity can be high even for signal-blind outputs.

\begin{wraptable}{r}{0.56\textwidth}
\vspace{-1em}
\centering
\small
\caption{\textbf{Embedding-metric diagnostic.}}
\label{tab:metric_illusion}
\vspace{-0.4em}

\resizebox{0.56\textwidth}{!}{
\begin{tabular}{llcccc}
\toprule
\textbf{Prediction} & \textbf{Category} & \textbf{tokens} & \textbf{WER} & \textbf{CER} & \textbf{BERT-F1} \\
\midrule
``the the''        & repetition     & 2 & 0.953 & 0.920 & \textbf{0.868} \\
``the''            & function word  & 1 & 0.965 & 0.962 & 0.826 \\
``In the eye the'' & fragment       & 4 & 0.964 & 0.871 & 0.814 \\
``I don't know''   & generic filler & 3 & 0.997 & 0.894 & 0.800 \\
\bottomrule
\end{tabular}
}

\vspace{0.2em}

\begin{minipage}{0.54\textwidth}
\tiny
Fixed, signal-blind outputs can achieve high BERT-F1 despite near-maximal edit distances.
The largest value exceeds reported MEG-decoded BERT-F1 levels~\citep{jayalath2025unlocking}, showing that embedding similarity is unreliable as a stimulus-recovery indicator under low SNR.
\end{minipage}

\vspace{-1.0em}
\end{wraptable}

\begin{figure*}[!ht]
  \centering
  \vspace{-0.8em}
  \begin{minipage}[t]{0.49\textwidth}
    \centering
    \includegraphics[width=\linewidth]{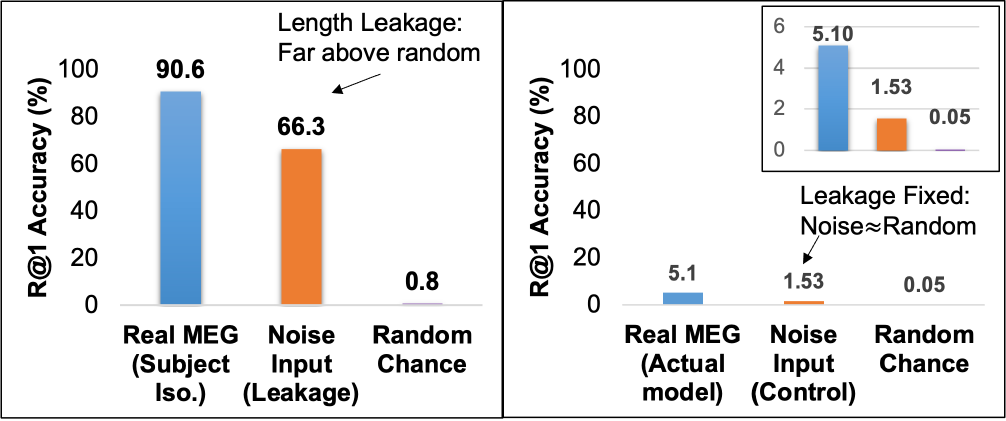}
    
    \vspace{0.15em}
    \footnotesize\textbf{(a) Length-leakage diagnostic.}
  \end{minipage}\hfill
  \begin{minipage}[t]{0.49\textwidth}
    \centering
    \includegraphics[width=\linewidth]{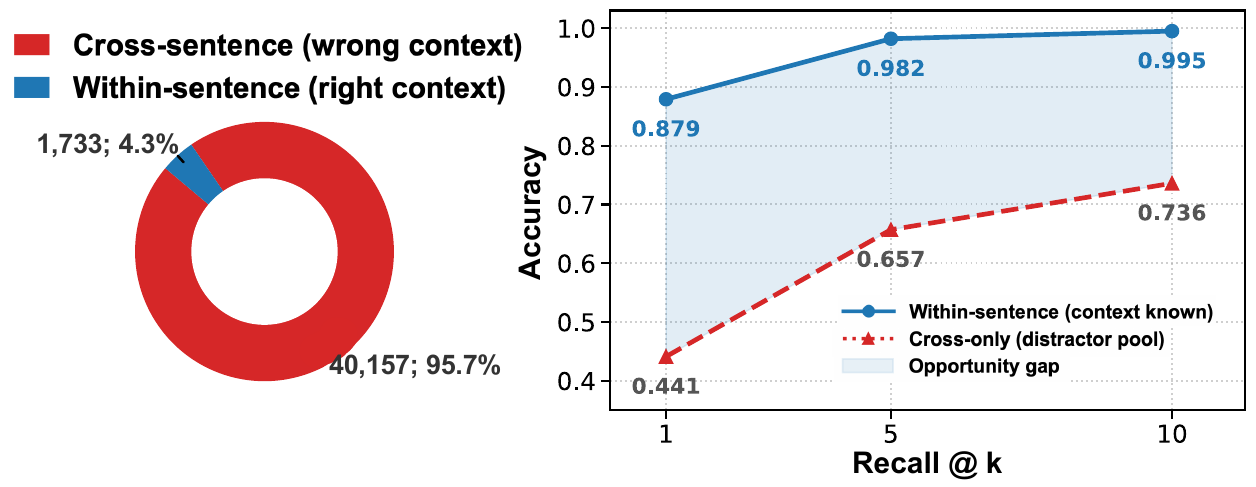}
    
    \footnotesize\textbf{(b) Cross-sentence bottleneck diagnostic.}
  \end{minipage}
  \caption{\textbf{Diagnostics for evaluation artefacts and contextual headroom.} (a) Under variable-length sentence inputs, duration cues support next-token ranking even with signal-blind neural input; fixed-length windows remove this shortcut, collapsing the noise baseline towards chance and revealing the weaker neural signal under leakage control. (b) Global closed-pool retrieval errors concentrate in cross-sentence competition rather than within-sentence ambiguity: restricting candidates to the ground-truth sentence bucket substantially raises Top-1 accuracy.}
  \label{fig:retrieval_bottleneck}
  \vspace{-1em}
\end{figure*}

\emph{Length leakage dominates variable-length next-token ranking.}
In the variable-length next-token diagnostic, padding patterns and attention masks can reveal sentence duration, correlating with sentence identity. Replacing MEG with Gaussian noise reaches \textbf{66.3\%} next-token R@1 (vs.\ \textbf{0.8\%} random); real MEG reaches \textbf{90.6\%}. Enforcing fixed-length inputs collapses both: real MEG falls to \textbf{5.10\%} and noise to \textbf{1.53\%}, near the fixed-length random baseline (\textbf{0.05\%}; Fig.~\ref{fig:retrieval_bottleneck}a). The variable-length advantage is therefore largely a duration shortcut rather than stimulus-evoked evidence.

\emph{Embedding-based metrics reward signal-blind outputs.}
Sentence-level decoding is often summarised by embedding-based similarity (e.g., BERT-F1), but these metrics quantify similarity in a pretrained language space and reward outputs that are fluent or generic regardless of stimulus content. Table~\ref{tab:metric_illusion} shows that fixed, signal-blind predictions such as ``the the'' or ``I don't know'' obtain BERT-F1 between 0.80 and 0.87 despite WER above 0.95; the highest of these (0.868) exceeds the reported MEG-decoded BERT-F1 level of approximately \(0.84\)~\citep{jayalath2025unlocking}. Free-form sentence-level decoding pipelines can also depend on automatically aligned audio--text supervision, which can drop negations or collapse clauses into repetitive loops (Appendix~\ref{app:align_noise}). Audio-segment retrieval avoids both issues: chance is explicit, the target is a unique audio window, and rank-based metrics cannot be inflated by these factors.

\textbf{Source 2: controlled window-level retrieval recovers local evidence but leaves measurable headroom for context.}
Under leakage-controlled retrieval, the pre-GCB scores in Table~\ref{tab:gcb_main_zero_shot} estimate the local stimulus-locked evidence source before any contextual aggregation is applied.
At this stage, duration cues, stimulus-identity overlap, candidate construction, and sentence-level aggregation have already been controlled.
Across all settings, Dense-TCNN yields stronger window-level MEG--audio discriminability than the receptive-field-matched exp-dilated reference. On Gwilliams zero-shot, the Dense-TCNN window-level baseline reaches R@1$=44.4\%$, showing that fixed-duration retrieval preserves measurable stimulus-locked discriminability rather than collapsing to chance.

The remaining errors identify where extra-window information can enter.
Fig.~\ref{fig:retrieval_bottleneck}b shows that 95.7\% of Top-1 failures select a candidate from the wrong sentence, while only 4.3\% select an incorrect candidate from the ground-truth sentence.
The oracle within-bucket diagnostic gives the corresponding controlled check: with the model and logits held fixed, restricting ranking to the ground-truth sentence raises R@1 from 0.44 to 0.88.
Thus, after leakage control, the local model already carries substantial within-sentence discriminative information, but full-pool ranking is still dominated by sentence-level competition.
This motivates treating context as a separate source: cross-window consistency can act as a soft sentence-bucket reweighting mechanism while leaving the full candidate pool intact.

\textbf{Source 3: score-space aggregation exposes recoverable contextual gain under frozen local logits.}

Given this diagnosis, we use the Base--GCB contrast in Table~\ref{tab:gcb_main_zero_shot} to quantify the recoverable contextual gain with the fixed GCB setting. With the encoder, candidate pool, and local logits fixed, the only change is a score-space aggregation of cross-window sentence-consistent evidence. On Gwilliams zero-shot, this contrast shifts Dense-TCNN R@1 from $44.4\%$ to $52.1\%$.

\providecommand{\sd}[2]{\ensuremath{#1{\scriptstyle\pm}#2}}
\providecommand{\bsd}[2]{\ensuremath{\mathbf{#1}{\scriptstyle\pm}\mathbf{#2}}}
\providecommand{\bn}[1]{\ensuremath{\mathbf{#1}}}
\providecommand{\ci}[2]{\ensuremath{[#1,#2]}}

\begin{table*}[!ht]
  \centering
  \setlength{\tabcolsep}{2.2pt}
  \renewcommand{\arraystretch}{0.98}
  \scriptsize
  \captionsetup{justification=justified,singlelinecheck=false}
  \caption{\textbf{GCB-over-Base contrast for auditing cross-window contextual aggregation under zero-shot stimulus-identity splits.}
Each backbone is evaluated before and after applying GCB to the same frozen retrieval logits, so the contrast estimates the recoverable contextual contribution while holding the encoder, candidate pool, and local logits fixed.
Dense denotes Dense-TCNN; Exp-dilated $(L{=}5)$ is the receptive-field-matched exp-dilated reference of~\citet{defossez2023decoding}; Exp-dilated $(L{=}10)$ is the deeper exp-dilated reference-depth variant.
Arrows indicate the preferred direction.
R@1, MRR, and R@10 are reported as percentages; $\Delta$ columns report absolute percentage-point gains.
R@1, MRR, R@10, and $\Delta$ are mean$\pm$std over three seeds, while MedR is the mean over three seeds.
Bold: the best value within each dataset block and column.}
  \label{tab:gcb_main_zero_shot}

  \resizebox{0.99\textwidth}{!}{%
  \begin{tabular}{@{}llcccccccccc@{}}
    \toprule
    \multirow{2}{*}{\textbf{Setting}} &
    \multirow{2}{*}{\textbf{Backbone}} &
    \multicolumn{4}{c}{\textbf{Local evidence}} &
    \multicolumn{4}{c}{\textbf{GCB-audited evidence}} &
    \multicolumn{2}{c}{\textbf{Attribution contrast}} \\
    \cmidrule(lr){3-6}
    \cmidrule(lr){7-10}
    \cmidrule(l){11-12}
    & &
    \textbf{R@1 (\%)$\uparrow$} & \textbf{MRR (\%)$\uparrow$} & \textbf{R@10 (\%)$\uparrow$} & \textbf{MedR$\downarrow$} &
    \textbf{R@1 (\%)$\uparrow$} & \textbf{MRR (\%)$\uparrow$} & \textbf{R@10 (\%)$\uparrow$} & \textbf{MedR$\downarrow$} &
    \textbf{$\Delta$R@1 (pp)$\uparrow$} & \textbf{$\Delta$MRR (pp)$\uparrow$} \\
    \midrule

    \multirow{3}{*}{\makecell[l]{\textbf{Gwilliams}\\$N{=}1{,}464$}}
      & Dense-TCNN
      & \bsd{44.4}{0.3} & \bsd{54.4}{0.2} & \bsd{73.6}{0.1} & \bn{2.0}
      & \bsd{52.1}{0.6} & \bsd{60.8}{0.5} & \bsd{77.3}{0.3} & \bn{1.0}
      & \sd{7.7}{0.9}\ensuremath{^{\dagger}} & \sd{6.4}{0.7}\ensuremath{^{\dagger}} \\

      & Exp-dilated ($L=5$)
      & \sd{41.9}{0.9} & \sd{51.9}{0.9} & \sd{71.1}{0.7} & \bn{2.0}
      & \sd{50.6}{0.7} & \sd{59.2}{0.7} & \sd{75.4}{0.6} & 1.3
      & \bsd{8.7}{0.6}\ensuremath{^{\dagger}} & \bsd{7.4}{0.6}\ensuremath{^{\dagger}} \\

      & Exp-dilated ($L=10$)
      & \sd{39.6}{0.3} & \sd{49.7}{0.2} & \sd{69.2}{0.1} & 3.0
      & \sd{46.2}{0.6} & \sd{55.3}{0.6} & \sd{72.7}{0.5} & 2.0
      & \sd{6.6}{0.8}\ensuremath{^{\dagger}} & \sd{5.7}{0.7}\ensuremath{^{\dagger}} \\
    \midrule

    \multirow{3}{*}{\makecell[l]{\textbf{MOUS}\\$N{=}825$}}
      & Dense-TCNN
      & \bsd{21.7}{0.4} & \bsd{30.2}{0.5} & \bsd{46.6}{0.5} & \bn{13.3}
      & \bsd{28.6}{0.6} & \bsd{37.0}{0.5} & \bsd{52.8}{0.3} & \bn{8.3}
      & \bsd{6.9}{0.2}\ensuremath{^{\dagger}} & \bsd{6.9}{0.1}\ensuremath{^{\dagger}} \\

      & Exp-dilated ($L=5$)
      & \sd{16.1}{0.8} & \sd{24.0}{0.7} & \sd{39.5}{1.0} & 21.7
      & \sd{21.5}{0.8} & \sd{29.3}{0.8} & \sd{44.0}{1.1} & 16.7
      & \sd{5.4}{0.1}\ensuremath{^{\dagger}} & \sd{5.3}{0.1}\ensuremath{^{\dagger}} \\

      & Exp-dilated ($L=10$)
      & \sd{13.4}{0.9} & \sd{20.8}{1.1} & \sd{35.1}{1.3} & 30.0
      & \sd{17.9}{1.2} & \sd{25.2}{1.3} & \sd{38.8}{1.6} & 24.7
      & \sd{4.5}{0.3}\ensuremath{^{\dagger}} & \sd{4.4}{0.2}\ensuremath{^{\dagger}} \\
    \bottomrule
  \end{tabular}%
  }

  \vspace{2pt}
  \begin{minipage}{0.99\textwidth}
  \tiny
  \(\dagger\)On $\Delta$ columns, the dagger indicates that the corresponding improvement is significant in all seed-specific paired sentence-cluster bootstrap tests at \(p<0.05\); full intervals are reported in Table~\ref{tab:zeroshot_seed_bootstrap}.
  \end{minipage}
\vspace{-2em}
\end{table*}

The same fixed audit setting yields a positive MOUS contrast, indicating that the contextual-source diagnostic is not specific to the Gwilliams evaluation setting.
The oracle bucket diagnostic is an upper-bound localisation tool, not the target achieved by GCB. GCB recovers only the part of this headroom already expressed as cross-window support in the frozen local logits, so it mainly corrects near-miss cross-sentence errors and cannot repair bucket-identification failures.
GCB instantiates this source conservatively: it uses the retrieval model's own cross-window evidence to apply a bucket-wise soft correction, while preserving the full candidate pool and within-bucket ordering.
Appendix~\ref{app:oracle_gap_decomp} analyses the remaining oracle--GCB gap in terms of bucket-identification failures and conservative correction under full-pool competition.
Together, the oracle diagnostic and the Base--GCB contrast separate two quantities: the headroom exposed by removing sentence-level competition, and the part of that headroom recoverable by auditable contextual reweighting.

\textbf{Component ablations.}
Table~\ref{tab:gcb_ablation} dissects GCB using a fixed Dense-TCNN retrieval model and its frozen Gwilliams zero-shot logits, so that each variant is compared under identical local evidence.
Removing cross-window pooling (GCB-single, $G=\{b\}$) drops R@1 to 0.423, slightly below the corresponding fixed-logit baseline (0.441), indicating that bucket support estimated from a single window is too unstable to provide reliable correction; restoring pooling adds $+0.092$ R@1 and $+0.076$ MRR (sentence-level paired bootstrap, $p<0.001$).
Removing the quantile gate drops R@1 to 0.378 --- aggregation must be driven by selective high-evidence logits rather than all top-$K$ scores.
Removing bucket-size normalisation inflates R@1 to 0.712, but this inflation correlates with bucket size (Spearman $r=0.27$, $p<0.001$).
Without bucket-size normalisation, pooled bucket support grows with the number of candidates in a sentence bucket, making the correction sensitive to sentence length.
This illustrates a broader risk for introducing contextual information to neural decoding: \textbf{without controlling the amount of contextual evidence, structural properties such as sequence length may dominate over content as retrieval cues, obscuring whether decoding actually relies on neural evidence for stimulus content.}

\begin{wraptable}[24]{r}{0.58\textwidth}
\centering
\vspace{-1.3em}
\scriptsize
\setlength{\tabcolsep}{2pt}
\renewcommand{\arraystretch}{1.08}
\captionsetup{justification=justified,singlelinecheck=false}

\caption{\textbf{GCB component ablations and pruning comparison on Gwilliams zero-shot.}
All variants are evaluated on the same fixed Dense-TCNN retrieval logits to isolate the effect of each GCB component.
GCB-single removes only cross-window pooling by replacing each sentence group with a singleton \(G=\{b\}\).
The w/o norm variant is diagnostic because it introduces sentence-length bias.
Hard pruning keeps only the top-supported sentence buckets and removes all other candidates from ranking.
Retrieval rates are reported as percentages; MedR is reported in rank units.
Bold marks the best value among controlled variants, excluding the diagnostic w/o norm row.}
\vspace{-0.6em}
\label{tab:gcb_ablation}

\resizebox{0.58\textwidth}{!}{
\begin{tabular}{lccccc}
\toprule
\textbf{Method} & \textbf{R@1 (\%)} & \textbf{R@5 (\%)} & \textbf{R@10 (\%)} & \textbf{MRR (\%)} & \textbf{MedR} \\
\midrule
\multicolumn{6}{l}{\textit{Component ablations}} \\
Base model only                         & 44.1 & 65.7 & 73.6 & 54.2 & 2 \\
w/o cross-window                        & 42.3 & 65.0 & 73.4 & 52.8 & 2 \\
w/o gate                                & 37.8 & 64.0 & 72.8 & 49.8 & 3 \\
GCB full                                & \textbf{52.1} & \textbf{71.4} & 77.7 & \textbf{61.4} & \textbf{1} \\
w/o norm$^{\dagger}$                    & 71.2 & 85.6 & 88.5 & 77.9 & 1 \\
\midrule
\multicolumn{6}{l}{\textit{Pruning comparison} \((S{=}3)\)} \\
Hard pruning                            & 50.5 & 62.3 & \textbf{82.3} & 58.2 & \textbf{1} \\
\bottomrule
\end{tabular}
}

\begin{minipage}{0.56\textwidth}
\tiny
\(^{\dagger}\)Bucket-size normalisation controls the quantity of sentence-level evidence: without it, pooled bucket support scales with sentence length.
The inflated R@1 of 71.2\% and its positive correlation with bucket size (\(r{=}0.27\)) show that, when this control is removed, length structure can dominate the apparent contextual gain.
Restoring cross-window pooling improves R@1 by \(+9.8\) pp and MRR by \(+8.6\) pp relative to GCB-single under sentence-level paired bootstrap resampling.
\end{minipage}

\vspace{-1em}
\end{wraptable}

\textbf{GCB vs.\ hard pruning.}
Restricting retrieval to the top-supported buckets (logits $\to-\infty$ elsewhere) increases R@10 by concentrating recall, but underperforms full GCB on R@1 and MRR. This separates the contextual effect from hard candidate-pool reduction: GCB acts through soft full-pool reweighting based on cross-window evidence.

\vspace{-1em}
\subsection{Audit controls}
\label{subsec:res_additional}
\vspace{-0.5em}

We next test when the contextual effect exposed by the Base--GCB contrast is interpretable.
The component and pruning controls in Table~\ref{tab:gcb_ablation} isolate the roles of cross-window pooling, selective evidence extraction, bucket-size normalisation, and soft full-pool reweighting.
Here we add two source-level controls: grouping perturbations test whether the effect depends on coherent sentence structure, and evidence-limited controls test whether it remains observable when local stimulus-locked evidence is weak.

\begin{wrapfigure}[16]{r}{0.7\textwidth}
\centering
\includegraphics[width=0.7\textwidth]{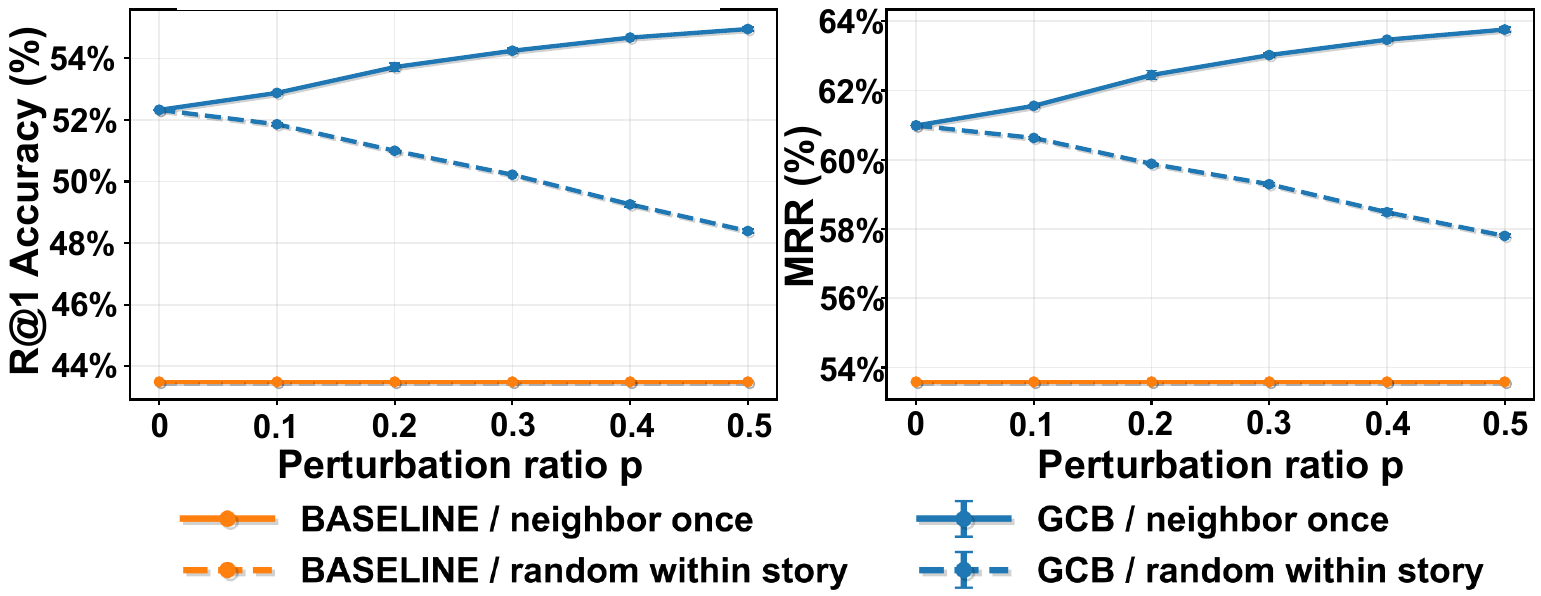}
\caption{
\textbf{Grouping perturbation robustness.}
Baseline retrieval is invariant to perturbations.
With GCB enabled, neighbour\_once perturbations can slightly improve performance, whereas random\_within\_story perturbations monotonically reduce performance as \(p\) increases.
}
\label{fig:group_robustness}
\vspace{-1em}
\end{wrapfigure}

\textbf{The contextual effect requires coherent sentence structure.}
We perturb query-to-sentence assignments while keeping the baseline logits fixed.
As expected, baseline retrieval is invariant.
With GCB, neighbour-boundary jitter slightly improves performance, consistent with mild annotation noise and local mixing acting as denoising, whereas random within-story reassignment monotonically degrades performance as \(p\) increases (Fig.~\ref{fig:group_robustness}).
Thus, coherent cross-window structure is necessary for the measured contextual effect.

\textbf{The contextual effect depends on local evidence.}
The Base--GCB contrast quantifies how much contextual structure can be recovered from existing stimulus-locked retrieval evidence.
We progressively attenuate local evidence while holding the encoder, candidate pool, sentence grouping, and GCB hyperparameters fixed.
Each query window \(x_w\) is mixed with a stimulus-mismatched real MEG window \(\tilde{x}_w\) from the same evaluation set, with \(\alpha\in[0,1]\) controlling retained target-specific evidence; the formula and surrogate variants are given in Appendix~\ref{app:evidence_attenuation}.
As \(\alpha\) decreases, \(\Delta\)R@1 falls from \(+8.6\) pp on clean MEG to \(+6.6\), \(+2.7\), \(+0.3\), and approximately \(0.0\) pp at \(\alpha{=}0.75,0.50,0.25,\leq0.10\), respectively; logit-space and phase/covariance surrogates show the same pattern.
Thus, score-space aggregation is separable from encoder learning, but becomes measurable only when local stimulus-locked evidence is present.

Brennan EEG provides an out-of-distribution evidence-limited counterpart: with near-chance local retrieval (R@1 \(<0.01\)), GCB does not yield a positive Base--GCB contrast, including under reduced candidate-pool sizes (Appendix~\ref{app:brennan_pool_sweep}).
Together, the controlled attenuation and EEG stress test show that contextual aggregation amplifies local evidence rather than replacing it.
The added runtime is small in absolute terms, approximately \(1.03\)\,ms per query (Appendix~\ref{app:opcount}).

\textbf{Additional operating-regime characterisations.}
Appendices~\ref{app:slr} and~\ref{app:flip_zeroshot} show that the contextual effect is bounded by frozen-logit quality: gains concentrate on near-miss corrections and contract when local evidence is weaker or cross-sentence competition is stronger.
Appendix~\ref{app:length_dependence} further characterises the length-dependent operating regime, with larger gains for short-to-medium sentences that provide consistent cross-window evidence.
We treat length dependence as an operating-regime characterisation rather than a causal estimate, since sentence length is coupled with bucket size and within-bucket difficulty.
Appendix~\ref{app:coupling} further shows that a training-time local--global coupling model does not improve over the local-only baseline in this low-SNR retrieval setting.

\section{Conclusions}
\label{sec:conclusion}

This work reframes non-invasive neural language decoding as an attribution problem: in low-SNR settings, apparent performance may reflect structural shortcuts, stimulus-locked evidence, or cross-window contextual aggregation, and final decoding scores alone do not reveal which source they reflect. Using stimulus-locked retrieval as a controlled endpoint, we show that variable-duration decoding can be dominated by a non-neural duration shortcut, whereas fixed-window retrieval under shortcut control preserves measurable MEG--audio evidence and exposes contextual headroom. The Base--GCB contrast audits this third source in score space: with the encoder, candidate pool, and local logits fixed, the observed gain estimates the contextual contribution recoverable under the fixed GCB setting. Perturbation and evidence-limited controls indicate that this contribution depends on coherent cross-window evidence and reliable stimulus-locked evidence, rather than grouping structure alone.

The controlled retrieval endpoint makes these sources separately observable, so the reported numerical estimates are endpoint-specific. Rather than making retrieval the final target, this endpoint is used to audit what information supports a decoding result. The same audit standard should apply to less structured pipelines: generative and end-to-end decoding pipelines should be evaluated with retrieval-style probes and task-specific controls over segmentation, context, candidate structure, decoder priors, and evaluation metrics. Without such audits, reported scores can show plausible outputs, but cannot establish what role stimulus-locked neural evidence plays in producing them.

\textbf{Convincing neural-language decoding should show not only what output was produced, but also to what extent that output is supported by neural evidence rather than by decoder priors, shortcuts, or other non-neural nuisances.}

\section*{Acknowledgments}
We would like to thank Auke Jan IJspeert  for their valuable help. The project was partially funded by the Swedish Research Council (Vetenskapsrådet) under award 2023-00493, and the NAISS under award 2025/22-1173, 2025/23-185, and 2026/3-376. For the purpose of open access, the author has applied a Creative Commons Attribution (CC BY) license to any Author Accepted Manuscript version arising from this submission.



\section*{Reproducibility Statement}

All experiments are based on existing public M/EEG datasets and publicly available pretrained audio models; Appendix~\ref{app:assets_ethics} reports the sources, licences, and access terms.
We provide the stimulus-identity splits, fixed-window construction, preprocessing, audio embeddings, encoder architectures, training hyperparameters, model selection, and compute setup in the Methods and Appendices~\ref{app:meta_manifest}--\ref{app:train_config}.
GCB is an inference-time post-processing aggregation in Sec.~\ref{subsec:mtd_context}, with implementation details, ablations, validation sweeps, and runtime measurements in Appendices~\ref{app:consensus}--\ref{app:opcount}.
We provide code for data preprocessing, model training, evaluation, ablations, and figure/table reproduction.

Raw data and pre-trained model weights are not redistributed and should be obtained from their original providers under the stated licences.

\section*{Use of Large Language Models (LLMs)}
We used LLMs (e.g., ChatGPT and Claude) to rephrase and polish the manuscript and to assist with coding tasks. All LLM-generated code was reviewed, edited, and integrated by the authors; the LLM did not design algorithms or produce experimental results.

\clearpage

\bibliography{neurips_2026}
\bibliographystyle{unsrtnat}

\clearpage
\appendix
\onecolumn

\section{Data, assets, and preprocessing}
\label{app:impl_overview}

\subsection{Sentence-level meta manifest and content-based splits}
\label{app:meta_manifest}

For each dataset, we construct a JSONL meta manifest that standardises sentence-level units across subjects and recordings.
Each row records: a unique sentence identifier, stimulus audio path, sentence onset/offset in stimulus time, and the corresponding MEG/EEG time interval in the original recording.
For Gwilliams and Brennan, sentence- and word-level timing information is taken from the released dataset event metadata.
For MOUS, sentence blocks are derived from BIDS \texttt{events.tsv} by identifying story vs.\ word-list segments and matching auditory sound events to their audio files; word-level timings are recovered from the stimulus audio using Whisper timestamps, as detailed in Appendix~\ref{app:mous_whisper}.

These timings are used as operational stimulus-side metadata for event-locked window construction.
They may contain residual alignment errors, which contribute to the shared alignment difficulty within each dataset because all retrieval baselines and GCB variants use the same timing metadata.

We define a \emph{stimulus identity} (content key) as the tuple (audio path, stimulus onset, stimulus offset).
All occurrences of the same content key, including repeated presentations across subjects or sessions, are assigned to the same training/validation/testing split.
Splits are constructed by sampling content keys with fixed ratios (70\%/10\%/20\%).
For Gwilliams, any training/validation windows that temporally overlap with testing windows on the same audio file are pruned, with the testing set taking priority, preventing cross-split temporal leakage.

\subsection{Existing assets, data access, and human-subject data}
\label{app:assets_ethics}
This work builds entirely on existing published resources. The three neural datasets---Gwilliams/MEG-MASC \citep{gwilliams2023introducing}, MOUS/Schoffelen \citep{schoffelen2019mous}, and Brennan \citep{brennan2019hierarchical}---are publicly available under open licences (CC0 1.0, RU-DI-HD-1.0, and CC-BY 4.0, respectively) and were accessed under their original repository terms. We do not collect new human-subject data, recruit participants, or redistribute raw neural recordings or stimulus audio; all participant consent, ethics approval, and risk procedures were the responsibility of the original collection studies. The pretrained models used as fixed components are wav2vec2-large-xlsr-53 \citep{baevski2020wav2vec} and Whisper large-v2 \citep{radford2023robust}, which are available under Apache-2.0 and MIT licences, respectively, and are not repackaged in any accompanying release.

\subsection{Code availability and reproducibility}
We provide anonymised code sufficient to reproduce the main analyses reported in this paper, covering manifest construction, stimulus-identity splitting, model training, logit computation, GCB and its ablations, and the main tables and figures. Raw neural recordings, stimulus audio, and third-party model weights are not included; these should be obtained directly from their original repositories under the terms described in Appendix~\ref{app:assets_ethics}. The README specifies the software environment, directory structure, and commands needed to replicate the experiments end-to-end.

\subsection{Neural preprocessing}
\label{app:preproc}

Neural windows are processed with a lightweight, shortcut-controlled recipe consistent across datasets and encoder variants:
(i) baseline correction using a short pre-onset interval,
(ii) robust scaling using median and interquartile range computed on the training split only, and
(iii) per-channel value clamping to reduce the influence of rare outliers.
We do not apply additional band-pass or notch filtering beyond dataset-provided preprocessing.
All subsequent modelling operates on tensors \(x \in \mathbb{R}^{C \times T}\) with \(T{=}360\) at 120\,Hz.
The corresponding high-level description is given in Methods, Sec.~\ref{sec:methods}.

\subsection{MOUS word-level alignment with Whisper}
\label{app:mous_whisper}

For MOUS, sentence blocks are obtained from BIDS \texttt{events.tsv} by identifying story vs.\ word-list segments.
Within each sentence block, word-level timings are recovered from the stimulus audio using Whisper (large-v2, language \texttt{nl}) with word timestamps enabled.
The resulting word onset times, expressed in stimulus coordinates, define the stimulus-locked neural windows used in the main experiments.

\paragraph{Estimated alignment error and its impact on retrieval.}
Whisper word-timestamp errors on naturalistic non-English speech are typically in the range of tens to a few hundred milliseconds, varying with speaking rate, audio quality, and language.
Our retrieval window is defined as \(X[t_w + \delta - 0.5,\; t_w + \delta + 2.5]\) (Eq.~\ref{eq:window_def}) at \(\delta=150\)\,ms, with a \(0.5\)\,s pre-onset margin and a \(2.5\)\,s post-onset coverage.
An onset jitter of \(\pm 200\)\,ms therefore shifts the window by roughly 7\% of its length, and the post-onset stimulus content remains inside the window.
We expect Whisper alignment noise to slightly degrade absolute MOUS retrieval performance relative to a hypothetical perfectly aligned ceiling, but not to alter the relative comparisons or the cross-source attribution conclusions in the main text: all baselines and GCB variants on MOUS use the same alignment, so any alignment-induced jitter is shared.

The cross-dataset transfer result (Gwilliams \(\rightarrow\) MOUS without retuning) is consistent with this interpretation: GCB hyperparameters fixed on Gwilliams still yield a positive contextual gain despite the noisier MOUS alignment.
Prior delay-sensitivity analysis in the same retrieval family also found that 3\,s M/EEG-to-speech retrieval is not hypersensitive to moderate global offsets: varying the M/EEG--speech delay from 0 to 200\,ms on Gwilliams2022 changed top-10 accuracy only modestly, with larger degradation appearing at 250--300\,ms~\citep{defossez2023decoding}.
A global delay sweep is not identical to word-level timestamp jitter, but it supports treating MOUS alignment noise primarily as an absolute-performance factor rather than as the driver of relative GCB gains.

\subsection{Audio embedding configuration}
\label{app:audio_embed_details}

Audio representations are obtained from a frozen wav2vec~2.0 encoder. We use \textbf{wav2vec2-large-xlsr-53}.
For each 3\,s audio window, we extract hidden activations from layers 14--18, average across layers, and retain the time-resolved representation.
We interpolate or resample along time to match the evaluation temporal resolution \(T{=}360\), yielding a \(D{\times}T\) target with \(D{=}1024\).
Standard waveform preprocessing, including resampling to 16\,kHz and waveform normalisation, is applied before feeding audio to wav2vec.
The same configuration is used for Gwilliams, MOUS, and Brennan so that performance differences arise from the neural side rather than changes in the audio representation.

\section{Retrieval model implementation details}

\subsection{Neural encoder families and instantiations}
\label{app:arch}

\textbf{Shared coordinate-conditioned spatial mixing.}
All encoder families share the same coordinate-conditioned spatial mixing frontend following prior MEG-to-audio retrieval work~\citep{defossez2023decoding}.
This module maps variable M/EEG sensor layouts to a fixed set of \(S{=}270\) virtual spatial channels, enabling the same temporal backbone design to operate across datasets with different native channel counts.
Sensor coordinates are embedded with 2D Fourier features and passed through a small MLP to produce sensor-to-virtual mixing weights, which are applied to the sensor\(\times\)time window to obtain a \(270\times T\) latent grid.

\textbf{Exp-dilated CNN backbone.}
The exp-dilated CNN backbone is adapted from~\citet{defossez2023decoding}.
We instantiate a 5-block exp-dilated CNN with kernel size 3 and dilations \([1,2,4,8,16]\), chosen to facilitate a controlled comparison against Dense-TCNN under matched effective receptive fields on 3\,s windows.
Each DilatedGLU block applies a dilated convolution with GLU gating and residual connections.
The temporal hidden width is \(d{=}320\), followed by a \(1{\times}1\) projection to the shared embedding dimension \(D{=}1024\).

\textbf{Dense-TCNN backbone.}
Dense-TCNN keeps the same widths (\(d{=}320\), projection to \(D{=}1024\)) but replaces the exponential dilation schedule with five residual blocks, each containing two dilated convolutions with moderate fixed dilations \(\{2,4\}\).
This yields a comparable effective temporal span to the exp-dilated backbone while promoting repeated short- and mid-range mixing with short residual paths.

\textbf{Concrete instantiations.}
Table~\ref{tab:arch_variants} summarises the concrete instantiations used in all main experiments.
All models share the same coordinate-conditioned spatial mixing frontend; only the temporal dilation pattern differs.
The number of native input sensors \(C\) depends on the dataset (\(C{=}208/272\) for MEG, \(C{=}60\) for EEG).

\begin{table}[!ht]
\centering
\small
\setlength{\tabcolsep}{6pt}
\caption{\textbf{Instantiated encoder families used for MEG and EEG retrieval}.
All models share the same coordinate-conditioned spatial mixing frontend; only the temporal dilation pattern differs.}
\renewcommand{\arraystretch}{1.15}
\begin{tabular}{lcccc}
\toprule
\textbf{Backbone} &
\textbf{\# Blocks} &
\textbf{Kernel} &
\textbf{Dilations} &
\textbf{Hidden / proj.} \\
\midrule
Dense-TCNN &
5 residual (2-layer) blocks &
3 &
\([2,4]\times 5\) &
\(d{=}320 \rightarrow D{=}1024\) \\
Exp-dilated CNN~\citep{defossez2023decoding} &
5 DilatedGLU blocks &
3 &
\([1,2,4,8,16]\) &
\(d{=}320 \rightarrow D{=}1024\) \\
\bottomrule
\end{tabular}
\label{tab:arch_variants}
\end{table}

\textbf{Receptive-field matching for a controlled comparison.}
\label{app:arch_rf_match}
We match effective receptive fields between the two temporal backbones for kernel size \(k{=}3\).
For a stack of dilated convolutions without striding, the receptive field is
\(R = 1 + \sum_i (k{-}1)d_i = 1 + 2\sum_i d_i\).
The 5-block exp-dilated CNN uses dilations \([1,2,4,8,16]\), so \(\sum_i d_i{=}31\) and \(R{=}63\) time steps.
Dense-TCNN contains five blocks with two dilated convolutions per block at dilations \(\{2,4\}\), giving \(\sum_i d_i{=}5(2{+}4){=}30\) and \(R{=}61\).
Thus both models operate on a comparable temporal context, and differences are not explained by access to a longer receptive field.

\textbf{Exp-dilated depth variant.}
\label{app:exp_d10_check}
We also report a deeper \(L{=}10\) exp-dilated variant that is closer to the original reference depth of~\citet{defossez2023decoding}.
This variant uses the same preprocessing, audio targets, retrieval loss, optimiser, model-selection protocol, and GCB post-processing as the other backbones; only the temporal depth is changed.
As shown in Table~\ref{tab:gcb_main_zero_shot}, the \(L{=}10\) exp-dilated variant does not improve over the receptive-field-matched \(L{=}5\) variant on either Gwilliams or MOUS.
We therefore treat the \(L{=}5\) model as the controlled receptive-field-matched reference, while the \(L{=}10\) rows document that the observed Dense-TCNN advantage is not explained by omitting a deeper exp-dilated instantiation.

\textbf{EEG instantiation.}
For Brennan EEG (\(C{=}60\)), we reuse the same spatial mixing frontend and temporal backbone families with identical hyperparameters; the only modality-specific change is the native input channel count.
Despite reusing the same encoder families, EEG window-level retrieval remains close to chance under shortcut-controlled evaluation, indicating that the limiting factor is neural evidence quality rather than architecture selection.

\subsection{Training hyperparameters}
\label{app:train}

Each neural encoder consists of a sensor-aware spatial projection, a subject-specific layer, and a temporal convolutional backbone.
It maps each neural input window to a \(1024\times T\) representation.
The audio encoder is kept frozen throughout training, so optimisation only updates the neural-side model.

Scores are computed without temporal pooling.
The neural representation is used directly, while each candidate audio representation is Frobenius \(L_2\)-normalised.
For a neural representation \(Z_i\) and an audio candidate representation \(A_c\), the retrieval logit is calculated using Equation~\ref{eq:retrieval_logits}.

The neural encoder is trained with a one-directional neural-to-audio contrastive retrieval objective using in-batch audio candidates as negatives.
Optimisation uses Adam with learning rate \(3{\times}10^{-4}\), zero weight decay, and effective batch size 256.
We use a linear warm-up over the first \(10\%\) of optimisation steps followed by cosine decay.
Training is run for at most 100 epochs, with early stopping patience 15 based on validation loss.
Gradients are clipped at a global norm of 1.0.

\subsection{Compute and evaluation settings}
\label{app:train_config}

Neural-model training experiments are run with PyTorch on a single NVIDIA A100-40GB GPU, and all reported training runs fit within the 40GB GPU memory.
The reported training runs use full precision with automatic mixed precision disabled and TF32 disabled.
The base retrieval models typically train within about 3 hours per run.
The end-to-end local--global coupling control and the Whisper-based duration diagnostic run within 24 hours on the same GPU.

Runtime benchmarks report mean \(\pm\) std over 5 runs, excluding one warm-up run (Table~\ref{tab:runtime_overhead}).

GCB aggregation, ablations, grouping perturbations and evidence-attenuation analyses are applied to precomputed frozen logits; these CPU post-processing analyses typically complete within 30 minutes per run.

The full research project required additional compute beyond the final reported runs, mainly for validation sweeps, debugging runs, and preliminary model variants.

\section{Structural and text-side diagnostic details}

\subsection{Whisper-based next-token diagnostic implementation details}
\label{app:gen_details}

We use a Whisper-conditioned next-token diagnostic to test whether sentence-level inputs expose duration cues.
This diagnostic follows the Whisper-based modelling setup of \citet{li2025brainecho} at the representation and token-scoring level, but it is not used as a free-form generative decoder.
Concretely, we train a Transformer-based VQ-VAE to map M/EEG features into a sequence compatible with the Whisper encoder input space, and evaluate the resulting representations through next-token ranking.
The reported metric is therefore token-level ranking accuracy rather than transcript generation quality.

To isolate the effect of length cues, we compare two input paradigms.
In the length-leakage-prone setting, we use sentence-aligned windows with natural boundaries, pad them to a common length, and mask valid time steps.
This protocol preserves information about the original sentence duration through the padding and mask structure.
In the shortcut-controlled setting, we split each sentence into fixed-length 3\,s windows and evaluate fixed-duration windows without direct access to variable sentence duration.
The resulting comparison is reported in Fig.~\ref{fig:retrieval_bottleneck}a.

\subsection{Fixed-window Gaussian-noise retrieval sanity check}
\label{app:fixed_window_noise_sanity}

The Whisper-based next-token diagnostic in Appendix~\ref{app:gen_details} is used to expose a variable-duration shortcut; it is not a retrieval baseline.
To verify that the main fixed-window retrieval endpoint does not retain the same signal-blind shortcut, we additionally run a retrieval-side sanity check.
Using the trained Dense-TCNN checkpoint on Gwilliams zero-shot, we replace each test M/EEG input window with signal-blind Gaussian noise while keeping the trained encoder, subject metadata, sensor coordinates, audio candidate pool, stimulus-identity split, and evaluation code fixed.
We evaluate both the local retrieval baseline and the same GCB post-processing rule.

Table~\ref{tab:fixed_window_noise_sanity} shows that fixed-window retrieval collapses to chance under Gaussian input.
For the standard Gaussian input, Base R@1 is \(0.00079\), close to the random baseline \(1/N=0.00068\), and GCB does not improve retrieval.
A per-window mean/std-matched Gaussian variant shows the same pattern.
Thus, under the fixed-window retrieval endpoint, signal-blind inputs do not produce above-chance retrieval, and sentence-level GCB grouping does not create a positive contrast without local stimulus-locked evidence.

\begin{table}[!ht]
\centering
\small
\setlength{\tabcolsep}{6pt}
\renewcommand{\arraystretch}{1.12}
\caption{\textbf{Fixed-window Gaussian-noise retrieval result on Gwilliams zero-shot.}
The trained base model is kept fixed, and each test M/EEG window is replaced by signal-blind Gaussian noise.
Chance is \(0.068\%\) for R@1, \(0.342\%\) for R@5, and \(0.683\%\) for R@10 with \(N=1{,}464\).}
\label{tab:fixed_window_noise_sanity}
\begin{tabular}{lcccc}
\toprule
\textbf{Input} & \textbf{Method} & \textbf{R@1 (\%)} & \textbf{R@5 (\%)} & \textbf{R@10 (\%)} \\
\midrule
Gaussian \(N(0,1)\)
& Base
& 0.079 & 0.335 & 0.700 \\
Gaussian \(N(0,1)\)
& +GCB
& 0.063 & 0.316 & 0.669 \\
\midrule
Mean/std-matched Gaussian
& Base
& 0.054 & 0.321 & 0.676 \\
Mean/std-matched Gaussian
& +GCB
& 0.060 & 0.311 & 0.673 \\
\midrule
Random chance
& --
& 0.068 & 0.342 & 0.683 \\
\bottomrule
\end{tabular}
\end{table}

\subsection{Limitations of embedding-based text similarity metrics}
\label{app:bertscore_limits}

Section~\ref{sec:experiments} (Table~\ref{tab:metric_illusion}) showed that fixed, signal-blind outputs obtain non-trivial BERT-F1 scores.
Here we provide additional discussion of why this happens.

Embedding-based text similarity metrics quantify similarity in a pretrained language representation space, which can reward outputs that are fluent, generic, or partially overlapping with the reference, even when they carry little stimulus-specific information.

Two properties matter in our diagnostic setting.
First, short, repetitive, or high-frequency fragments can receive non-trivial similarity scores because they are locally plausible and overlap with common contextual patterns.
Second, meaning-critical edits, such as negation flips or dropped content, can be under-penalised relative to exact-match measures.
Consequently, a fixed signal-blind baseline can obtain comparatively high embedding-based similarity while exhibiting near-maximal WER/CER.

For completeness, we therefore report embedding-based similarity alongside lexical metrics (WER/CER; BLEU), and interpret embedding scores only as coarse indicators of surface-level plausibility rather than evidence of neurally grounded decoding.
This motivates using an audio-based retrieval setting for primary evaluation, where chance is explicit and performance depends on matching stimulus-aligned audio candidates rather than producing text that is merely plausible under a language prior.

\subsection{Forced-alignment label noise in sentence-level generation}
\label{app:align_noise}

Free-form sentence-level decoding pipelines often require word- or sentence-level timestamps to form audio--text training targets, which are frequently obtained via automatic forced alignment.
When alignment fails, pseudo-labels can distort meaning, for example by dropping negation, or introduce degenerate repetitions, creating a mismatch between the true audio content and the training label.
This can penalise models even when predictions are consistent with the actual stimulus audio.
Table~\ref{tab:alignment_errors} shows an illustrative example where the forced-alignment transcript drops a critical negation and collapses a clause into a repetitive loop, motivating audio-based supervision and retrieval-based evaluation.

\begin{table}[!ht]
  \centering
  \small
  \renewcommand{\arraystretch}{1.3}
  \caption{\textbf{Illustrative label noise from forced alignment.}
  The automated transcript drops the negation ``didn't'' (inverting meaning) and collapses a clause into a repetitive loop.}
  \label{tab:alignment_errors}
  \begin{tabular}{p{0.95\linewidth}}
    \toprule
    \textbf{Actual audio content:} \\
    \textit{``...the short-eared people \underline{didn't} really know the secrets of the long ears and had pulled down all the moai statues and destroyed some of the tablets.''} \\
    \midrule
    \textbf{Force-aligned transcript (training label):} \\
    \textit{``...the short eared people really know the secrets \textbf{the the the}''} \\
    \bottomrule
  \end{tabular}
  \vspace{-2.0em}
\end{table}

\section{Group Context Bias implementation details}
\label{app:consensus}

\subsection{Implementation overview}
\label{app:consensus_overview}

Group Context Bias (GCB) is implemented as an inference-time post-processing layer applied to precomputed retrieval logits.
It does not modify the neural encoder, audio embeddings, candidate pool, or within-bucket candidate ordering.
The method operates on the same closed candidate pool used for baseline retrieval; each candidate is de-duplicated by stimulus identity and window index, so each candidate corresponds to a unique audio window.

We use the notation from Sec.~\ref{subsec:mtd_context}.
For base logits \(L\in\mathbb{R}^{B\times N}\), each candidate \(j\) is assigned to a sentence bucket \(c(j)\in\{1,\dots,M\}\), with bucket set
\[
\mathcal{B}_s=\{j\in[1,N]:c(j)=s\}.
\]
Query windows are grouped by the presented sentence, and \(G\subseteq\{1,\dots,B\}\) denotes one such query group.
GCB first extracts high-confidence candidate evidence from each query window, pools this evidence across \(G\) at the sentence-bucket level, and applies an additive bucket-wise logit correction to all candidates in the top-supported buckets.

\subsection{Algorithmic form of full GCB}
\label{app:consensus_algorithms}

Algorithm~\ref{alg:gcb_main} gives an implementation-equivalent form of full GCB.
The equations in Sec.~\ref{subsec:mtd_context} provide the formal definition; the algorithm is included to make the post-processing order explicit.

\begin{algorithm}[!ht]
\caption{Group Context Bias}
\label{alg:gcb_main}
\begin{algorithmic}[1]
\REQUIRE Base logits \(L\in\mathbb{R}^{B\times N}\);
candidate-to-bucket map \(c(j)\in\{1,\dots,M\}\);
bucket sets \(\{\mathcal{B}_s\}_{s=1}^M\);
query-window groups \(\mathcal{G}\);
hyperparameters \((K,q,m,S,\gamma_{\mathrm{gcb}})\);
bucket normaliser \(\mathrm{norm}(\cdot)\).
\ENSURE Corrected logits \(L^{\mathrm{GCB}}\in\mathbb{R}^{B\times N}\).

\STATE \(L^{\mathrm{GCB}}\gets L\)

\FOR{each query group \(G\in\mathcal{G}\)}
  \STATE Initialise empty evidence lists \(\{\mathcal{E}_s\}_{s=1}^M\)

  \FOR{each query window \(b\in G\)}
    \STATE \(\mathcal{I}_b \gets \mathrm{TopK}(L_{b,:},K)\)
    \STATE \(\tau_b \gets \mathrm{Quantile}_q(L_{b,:})\)

    \FOR{each candidate \(j\in\mathcal{I}_b\)}
      \IF{\(L_{b,j}\ge\tau_b\)}
        \STATE \(s\gets c(j)\)
        \STATE \(\mathcal{E}_s \gets \mathcal{E}_s\uplus\{L_{b,j}-\tau_b\}\)
      \ENDIF
    \ENDFOR
  \ENDFOR

  \FOR{each bucket \(s\in\{1,\dots,M\}\)}
    \IF{\(\mathcal{E}_s=\emptyset\)}
      \STATE \(U_G(s)\gets 0\)
    \ELSE
      \STATE \(U_G(s)\gets \mathrm{norm}(|\mathcal{B}_s|)\cdot\mathrm{MeanTopm}(\mathcal{E}_s,m)\)
    \ENDIF
  \ENDFOR

  \STATE \(\mathcal{S}_G\gets \mathrm{TopS}(\{U_G(s)\}_{s=1}^M,S)\)

  \FOR{each query window \(b\in G\)}
    \FOR{each bucket \(s\in\mathcal{S}_G\)}
      \FOR{each candidate \(j\in\mathcal{B}_s\)}
        \STATE \(L^{\mathrm{GCB}}_{b,j}\gets L^{\mathrm{GCB}}_{b,j}+\gamma_{\mathrm{gcb}}U_G(s)\)
      \ENDFOR
    \ENDFOR
  \ENDFOR
\ENDFOR

\STATE \textbf{return} \(L^{\mathrm{GCB}}\)
\end{algorithmic}
\end{algorithm}

\subsection{GCB-single ablation}
\label{app:gcb_single}

GCB-single is a component ablation that isolates the cross-window pooling step in Sec.~\ref{subsec:mtd_context}.
It keeps the candidate buckets, local evidence extraction, bucket-size normalisation, top-\(S\) bucket selection, and additive bucket-wise correction exactly as in Eqs.~\eqref{eq:gcb_bucket}--\eqref{eq:gcb_bias}.
The only change is the query-side grouping: instead of using sentence-level query groups, we replace them by the singleton partition
\begin{equation}
\mathfrak{G}_{\mathrm{single}}
=
\bigl\{\{b\}: b\in[1,B]\bigr\}.
\label{eq:gcb_single_groups}
\end{equation}

For each singleton group \(G=\{b\}\), the retained evidence \(\mathcal{E}_G(s)\), bucket support \(U_G(s)\), selected bucket set \(\mathcal{S}_G\), and corrected logits are computed using the same definitions as full GCB in Eqs.~\eqref{eq:gcb_Edef}--\eqref{eq:gcb_bias}.
We denote the resulting corrected logits by \(L^{\mathrm{single}}\) to distinguish this ablation from full sentence-level GCB.

Thus, GCB-single preserves all local score processing and bucket-wise correction operations, but prevents neighbouring windows from the same presented sentence from contributing to the bucket support \(U_G(s)\).
The comparison between GCB-single and full GCB therefore isolates the contribution of cross-window evidence pooling, rather than changes in gating, bucket-size normalisation, selected bucket count, bias scale, or candidate-pool construction.

\subsection{Hyperparameter selection and validation sweep}
\label{app:gcb_scan}

We select GCB hyperparameters on the Gwilliams validation split and keep the selected configuration fixed for all reported test evaluations, including the transfer to MOUS.
The fixed configuration used in the main experiments is
\(K{=}128\), \(q{=}0.95\), \(m{=}3\), \(S{=}3\), bucket-size square-root normalisation, and \(\gamma_{\mathrm{gcb}}{=}0.7\).
The GCB-single ablation uses the same values; the only change is that each query group is replaced by a singleton \(G=\{b\}\), so the comparison isolates cross-window evidence pooling rather than hyperparameter differences.
The gain parameter \(\gamma_{\mathrm{gcb}}\) is further characterised in Appendix~\ref{app:gamma_sweep}.

Table~\ref{tab:sweep_summary} reports representative validation configurations from the sweep rather than the full grid.
The main pattern is that score-based aggregation with bucket-size square-root normalisation is stable across nearby choices of \(K\), \(q\), and aggregation form.
By contrast, removing bucket-size normalisation or replacing it with count-based normalisation weakens or destabilises sentence-bucket recovery.
We also include rank- or count-only analogues that discard local-logit magnitude and retain only whether a bucket appears among high-ranked candidates.
These variants are less stable than score aggregation, which preserves the original magnitude, indicating that cross-window bucket support is better captured by the relative size of local evidence than by bucket hit frequency alone.
These validation checks support the use of GCB as an evidence-weighted score-space intervention.

\begin{table}[!ht]
\centering
\small
\caption{\textbf{Representative validation sweep configurations for GCB.}
The table reports selected configurations from the validation sweep rather than the full grid.
Score-based aggregation with \texttt{bucket\_sqrt} normalisation is stable across nearby settings, whereas unnormalised and count-based variants are weaker or less stable.
The final reported configuration uses \(K{=}128\), \(q{=}0.95\), \(m{=}3\), \(S{=}3\), \texttt{bucket\_sqrt}, and \(\gamma_{\mathrm{gcb}}{=}0.7\).}
\label{tab:sweep_summary}
\begin{tabular}{@{}llllllcc@{}}
\toprule
\(K\) & \(q\) & agg & \(m\) & norm & \(S\) & sent@1 (\%) \(\uparrow\) & sent@5 (\%) \(\uparrow\) \\
\midrule
64  & 0.95 & mean & 3 & bucket\_sqrt & 3 & 89.0 & 95.0 \\
128 & 0.95 & lse  & 1 & bucket\_sqrt & 3 & 90.0 & 96.0 \\
256 & 0.98 & mean & 3 & bucket\_sqrt & 3 & 89.0 & 94.0 \\
\midrule
128 & 0.95 & mean & 3 & none          & 3 & 82.0 & 87.0 \\
128 & 0.95 & mean & 3 & kept\_count   & 3 & 75.0 & 85.0 \\
128 & 0.95 & mean & 3 & bucket\_count & 3 & 21.0 & 35.0 \\
\bottomrule
\end{tabular}
\end{table}

\subsection{Complexity-only operation counts and runtime overhead}
\label{app:opcount}

To complement wall-time measurements, we report a complexity-only operation count in Table~\ref{tab:opcount} and an end-to-end runtime benchmark in Table~\ref{tab:runtime_overhead}.
Here, a multiply--accumulate operation (MAC) denotes one elementwise product and accumulation term in a dot product; under this convention, the similarity stage requires \(Q\cdot N\cdot D\cdot T\) MACs.
``GCB'' counts only explicit elementwise arithmetic and bucket writes in our Python implementation, excluding kernel-internal costs of \texttt{topk}/\texttt{quantile}/\texttt{unique} and other selection primitives.
As a result, the complexity-only counts should be interpreted as arithmetic lower bounds and are not expected to predict wall time precisely.

\begin{table}[!ht]
\centering
\small
\setlength{\tabcolsep}{6pt}
\renewcommand{\arraystretch}{1.15}
\caption{\textbf{Complexity-only count summary on Gwilliams zero-shot}
(\(Q{=}71{,}736\), \(N{=}1{,}464\), \(D{=}1024\), \(T{=}360\)).
Similarity MACs are exact by tensor shape, where one MAC denotes one multiply--accumulate term in the dot product.
GCB counts are implementation-level counters reported in their native units, including retained elements, unique buckets, explicit additions, and bucket writes; kernel-internal selection costs are excluded.}
\label{tab:opcount}
\begin{tabular}{lcc}
\toprule
\textbf{Counted quantity} & \textbf{Per query window} & \textbf{Total over test set} \\
\midrule
Similarity MACs (\(NDT\)) & \(5.40\times 10^{8}\) & \(3.87\times 10^{13}\) \\
\midrule
GCB: gated rank@\(K\) retained & 74.0 elems & \(5.31\times 10^{6}\) elems \\
GCB: unique sentence buckets & 8.80 buckets & \(6.31\times 10^{5}\) buckets \\
GCB: fuse add (\(N\) adds) & 1,464 adds & \(1.05\times 10^{8}\) adds \\
GCB: bucket writes & 1.33 writes & \(9.51\times 10^{4}\) writes \\
\bottomrule
\end{tabular}
\end{table}

\begin{table}[!ht]
  \centering
  \small
  \setlength{\tabcolsep}{7pt}
  \renewcommand{\arraystretch}{1.15}
  \caption{\textbf{End-to-end evaluation runtime on Gwilliams zero-shot} (\(N{=}1{,}464\); mean \(\pm\) std over 5 runs, 1 warm-up excluded).}
  \label{tab:runtime_overhead}
  \begin{tabular}{lccc}
    \toprule
    \textbf{Condition} & \textbf{Wall time (s)} & \textbf{ms/query} & \textbf{Overhead (\%)} \\
    \midrule
    Baseline & \(136.0 \pm 0.9\) & \(1.896 \pm 0.013\) & -- \\
    +GCB     & \(209.9 \pm 2.9\) & \(2.926 \pm 0.041\) & 54.3 \\
    \bottomrule
  \end{tabular}
\end{table}

\noindent\textbf{Note.}
The large gap between complexity-only counts and measured wall-time overhead is expected: selection primitives, kernel launches, and memory traffic can dominate runtime despite small explicit arithmetic.

\section{Robustness and mechanism analyses}
\label{app:mechanism}

This section provides additional analyses supporting the interpretation of GCB as an inference-time near-miss correction mechanism and as an auditable source-attribution intervention.
We first clarify the event-locked framework and the query-side and candidate-side grouping assumptions used by the retrieval benchmark.
We then analyse zero-shot rank flips, decompose the oracle--GCB headroom gap, and characterise the gain parameter \(\gamma_{\mathrm{gcb}}\) to show which part of the oracle ceiling is reachable by soft score-space aggregation and where more aggressive contextual weighting changes the nature of the intervention. Next, we test the same mechanism under the harder Gwilliams session-isolated setting, where the candidate pool is larger and local ranking is weaker.
We further characterise the operating regime of GCB through sentence-length dependence, within-MEG evidence attenuation, and Brennan EEG as an out-of-distribution evidence-limited stress test.
Finally, we report complementary controls and visualisations, including a training-time local--global coupling baseline, score-space visualisation of GCB's effect on representative queries, qualitative spatial-mixing visualisation, and token-level examples of pre/post-GCB candidate rescoring.
Throughout this section, the baseline denotes the corresponding backbone without inference-time aggregation, and GCB is applied to the same frozen retrieval logits.

\subsection{Event-locked framework and grouping assumptions}
\label{app:event_locked_grouping}

This work uses an event-locked stimulus-retrieval framework.
Each neural query window is extracted relative to a stimulus-side event time, and each candidate is a stimulus-aligned audio window from an explicit retrieval pool.
This follows the standard event-related logic of M/EEG analysis, where neural responses are analysed relative to externally defined events such as stimulus onsets, word onsets, or task events~\citep{luck2014introduction,beres2017time}.
Classic event-related potential/field (ERP/ERF) analyses use the same principle: P300-like responses are defined relative to task-relevant stimulus events~\citep{sutton1965evoked,polich2007updating}, and the N400 is a canonical example of word-locked language-related brain activity~\citep{kutas1980reading}.
Following this principle, we use stimulus-side timing metadata to align the neural windows.

This event-locked construction is important for the attribution question.
A dense sliding-window endpoint would introduce many neural segments that are not tied to a unique word-level or sentence-level target unless an additional segmentation or assignment rule is imposed.
It would therefore mix retrieval with activity detection, temporal segmentation, and target assignment.
It could also expose non-stimulus regularities from the experimental design or recording structure, because many windows would be defined by acquisition time rather than by a stimulus event.
By contrast, fixed event-locked windows give every query the same temporal support and a well-defined stimulus-aligned audio target, removing variable-duration cues and keeping the retrieval decision tied to stimulus-locked evidence.

The timing annotations may contain residual alignment noise.
We treat this as part of the shared alignment difficulty within each dataset: all compared methods use the same event times, so timing noise affects absolute alignment quality rather than giving a method-specific advantage.
The grouping perturbation analysis in Fig.~\ref{fig:group_robustness} provides an empirical check on the role of boundary structure: small neighbour-boundary perturbations do not destroy the contextual effect, whereas random within-story reassignment monotonically degrades GCB while leaving the baseline unchanged.

Within this event-locked framework, GCB uses sentence-level structure in two distinct ways: query-side grouping and candidate-side bucket structure.
On the query side, windows from the same presented sentence are grouped so that neighbouring observations can contribute to the same contextual estimate.
On the candidate side, audio candidates are partitioned into sentence buckets, allowing the pooled contextual support to be applied as a bucket-wise logit correction.
Thus, query-side grouping determines \emph{which observations are pooled}, whereas candidate-side grouping determines \emph{which candidates receive the shared contextual bias}.

Candidate-side buckets make the contextual source explicit in score space.
The local model first scores every query against the full candidate pool, fixing the window-level stimulus-locked evidence.
Sentence buckets are introduced only after local scoring, so that cross-window support can be applied as a bucket-wise correction while keeping local evidence and contextual aggregation separable for attribution.

This grouping structure is not part of the local retrieval model.
Local logits are computed over the full query--candidate product \(Q\times C\), and sentence identities enter only after local scoring, through the oracle diagnostic and the GCB correction.
The neural encoder, audio embeddings, candidate pool, and local logits are unchanged by grouping.
GCB also does not receive the ground-truth candidate bucket; it estimates supported buckets from the frozen local logits and selects only the top-\(S\) buckets for correction.
The oracle diagnostic is separate: it restricts retrieval to the ground-truth sentence bucket and is used only to measure the within-bucket ceiling.

The grouping perturbation experiment tests whether this structure provides coherent cross-window evidence rather than an oracle label.
Random grouping perturbations degrade GCB, showing that the gain depends on coherent cross-window structure rather than grouping metadata alone.
Evidence attenuation keeps the candidate pool, grouping structure, encoder, and GCB hyperparameters fixed while weakening target-specific local evidence; the GCB gain collapses as local retrieval approaches chance.
Thus, sentence grouping organises and amplifies locally supported evidence, but does not replace stimulus-locked evidence.

The oracle--GCB decomposition shows the same limitation quantitatively.
The ground-truth bucket is included in GCB's selected top-\(S\) buckets for only \(57.0\%\) of all queries and \(42.4\%\) of baseline errors.
If candidate-side grouping acted as an oracle label, the correct bucket would always be selected.
Instead, GCB must infer bucket support from noisy local logits, so the remaining oracle headroom is governed by bucket-identification failures, conservative soft correction under full-pool competition, and residual within-bucket ranking limits.

A less structured implementation would need to estimate stimulus events, temporal segments, and candidate groups rather than receive them as stimulus-side metadata.
That setting introduces additional sources of variation and is a broader decoding problem than the controlled attribution endpoint studied here.

\subsection{Zero-shot bootstrap and rank-flip diagnostics}
\label{app:flip_zeroshot}

Table~\ref{tab:zeroshot_seed_bootstrap} reports seed-specific paired sentence-cluster bootstrap tests for the zero-shot Base--GCB improvements in Table~\ref{tab:gcb_main_zero_shot}.
The main table reports absolute Base and +GCB metrics together with direct seed-wise point differences.
Here we report only paired bootstrap intervals for the improvement statistic, avoiding shifted absolute intervals that should not be interpreted as confidence intervals for the +GCB metric itself.
Across both datasets, all intervals are strictly positive and all seed-specific tests return \(p<0.05\), indicating that the Base--GCB contrast is stable across seeds and backbones.
\providecommand{\cipct}[2]{\ensuremath{[#1\%,\,#2\%]}}

\providecommand{\cipp}[2]{\ensuremath{[#1,\,#2]}}

\begin{table*}[!ht]
  \centering
  \footnotesize
  \setlength{\tabcolsep}{5pt}
  \renewcommand{\arraystretch}{1.05}
  \captionsetup{justification=justified,singlelinecheck=false}
  \caption{\textbf{Seed-specific paired sentence-cluster bootstrap tests for zero-shot GCB-over-Base improvements.}
Intervals are 95\% paired sentence-cluster bootstrap confidence intervals for the absolute improvement after applying GCB to the frozen Base logits.
All intervals are reported as absolute percentage-point differences.
All intervals are strictly positive, and all corresponding bootstrap tests return \(p<0.05\).
Only improvement intervals are reported.}
  \label{tab:zeroshot_seed_bootstrap}

  \begin{tabular}{@{}llcccc@{}}
    \toprule
    \multirow{2}{*}{\textbf{Backbone}} &
    \multirow{2}{*}{\textbf{Seed}} &
    \multicolumn{2}{c}{\textbf{Gwilliams}} &
    \multicolumn{2}{c}{\textbf{MOUS}} \\
    \cmidrule(lr){3-4}
    \cmidrule(l){5-6}
    & &
    \textbf{$\Delta$R@1 (pp)} & \textbf{$\Delta$MRR (pp)} &
    \textbf{$\Delta$R@1 (pp)} & \textbf{$\Delta$MRR (pp)} \\
    \midrule

    Dense-TCNN & 1 & \cipp{10.0}{11.6} & \cipp{8.3}{9.6} & \cipp{6.2}{7.2} & \cipp{6.4}{7.3} \\
    Dense-TCNN & 2 & \cipp{8.7}{10.4} & \cipp{7.3}{8.6} & \cipp{6.4}{7.5} & \cipp{6.5}{7.4} \\
    Dense-TCNN & 3 & \cipp{8.6}{10.4} & \cipp{7.1}{8.4} & \cipp{6.5}{7.5} & \cipp{6.4}{7.3} \\
    \midrule

    Exp-dilated $(L{=}5)$ & 1 & \cipp{10.5}{12.3} & \cipp{8.8}{10.1} & \cipp{5.0}{6.0} & \cipp{4.9}{5.8} \\
    Exp-dilated $(L{=}5)$ & 2 & \cipp{9.7}{11.4} & \cipp{8.0}{9.4} & \cipp{4.8}{5.7} & \cipp{4.8}{5.7} \\
    Exp-dilated $(L{=}5)$ & 3 & \cipp{10.5}{12.1} & \cipp{8.9}{10.2} & \cipp{5.0}{5.9} & \cipp{4.8}{5.7} \\
    \midrule

    Exp-dilated $(L{=}10)$ & 1 & \cipp{7.6}{9.6} & \cipp{6.4}{7.9} & \cipp{4.2}{5.2} & \cipp{4.1}{5.0} \\
    Exp-dilated $(L{=}10)$ & 2 & \cipp{8.4}{10.3} & \cipp{7.1}{8.6} & \cipp{3.7}{4.7} & \cipp{3.8}{4.7} \\
    Exp-dilated $(L{=}10)$ & 3 & \cipp{8.9}{10.7} & \cipp{7.7}{9.1} & \cipp{4.3}{5.1} & \cipp{4.2}{4.9} \\
    \bottomrule
  \end{tabular}
\end{table*}
We next characterise where the Top-1 changes come from on Gwilliams zero-shot using the fixed Dense-TCNN run used for the component diagnostics.
This analysis is not a separate aggregate performance estimate; it is an error-transition diagnostic on the same frozen local logits.
A bad-to-good flip denotes a query that is not Top-1 under the baseline but becomes Top-1 after GCB, whereas a good-to-bad flip denotes the reverse.
Across \(Q{=}71{,}736\) queries, GCB produces 6,826 bad-to-good flips and 484 good-to-bad regressions, yielding 6,342 net Top-1 corrections (Table~\ref{tab:flip_zeroshot}).
Thus, the Top-1 change is dominated by corrections rather than by a symmetric redistribution of correct and incorrect predictions.

\begin{table}[!ht]
\centering
\small
\setlength{\tabcolsep}{5pt}
\renewcommand{\arraystretch}{1.12}
\captionsetup{justification=justified,singlelinecheck=false}
\caption{\textbf{Top-1 rank flips on Gwilliams zero-shot} (\(Q{=}71{,}736\) queries; Dense-TCNN backbone).
Correction balance is defined as
\((\mathrm{bad{\rightarrow}good}-\mathrm{good{\rightarrow}bad})/
(\mathrm{bad{\rightarrow}good}+\mathrm{good{\rightarrow}bad})\).}
\label{tab:flip_zeroshot}
\vspace{0.4em}
\begin{tabular}{lrrrr}
\toprule
\textbf{Setting} &
\textbf{Good$\rightarrow$bad} &
\textbf{Bad$\rightarrow$good} &
\textbf{Net corr.} &
\textbf{Corr. balance (\%)} \\
\midrule
GCB over Base & 484 & 6,826 & +6,342 & 86.8 \\
\bottomrule
\end{tabular}
\end{table}

Table~\ref{tab:zeroshot_rank_bucket} further reports Top-1 correction rates stratified by the baseline rank bucket of the correct target.
GCB primarily recovers near misses: 9.3\% of rank 2--5 errors are promoted to Top-1, compared with 3.8\% for rank 6--10 and only 0.2\% for rank \(>10\).
This supports the intended interpretation of GCB as a conservative cross-window reweighting mechanism: it mainly acts when the correct target is already supported by the local logits, rather than rescuing arbitrarily low-ranked candidates.

\begin{table}[!ht]
\centering
\small
\setlength{\tabcolsep}{6pt}
\renewcommand{\arraystretch}{1.12}
\captionsetup{justification=justified,singlelinecheck=false}
\caption{\textbf{Top-1 correction rate by baseline target-rank bucket on Gwilliams zero-shot.}
Among baseline errors in each rank bucket, the table reports the fraction promoted to Top-1 by GCB.}
\label{tab:zeroshot_rank_bucket}
\vspace{0.4em}
\begin{tabular}{lccc}
\toprule
\textbf{Backbone} &
\textbf{Rank 2--5} &
\textbf{Rank 6--10} &
\textbf{Rank \(>10\)} \\
\midrule
Dense-TCNN & 9.3\% & 3.8\% & 0.2\% \\
\bottomrule
\end{tabular}
\end{table}

\subsection{Oracle--GCB headroom decomposition}
\label{app:oracle_gap_decomp}

The oracle sentence-bucket diagnostic and GCB answer different questions.
The oracle diagnostic restricts retrieval to the ground-truth sentence bucket while keeping the model, logits, and target candidate fixed.
It therefore exposes the maximum headroom available from removing cross-sentence distractors under unchanged within-bucket ordering.
GCB is deliberately weaker: it must infer supported buckets from the frozen local logits and then applies only a soft bucket-wise correction over the full candidate pool.

For the Gwilliams zero-shot Dense-TCNN component-analysis run, the local baseline reaches \(R@1=44.1\%\), while the oracle within-bucket diagnostic reaches \(R@1=87.9\%\), \(R@5=98.2\%\), and \(R@10=99.5\%\).
Thus, the oracle headroom is \(87.9\%-44.1\%=43.8\) percentage points.
GCB reaches \(R@1=52.1\%\), recovering \(52.1\%-44.1\%=8.0\) percentage points, or \(18.3\%\) of the oracle headroom.
This partial recovery is expected if GCB is interpreted as a conservative near-miss contextual correction rather than as an oracle approximation.

Table~\ref{tab:oracle_gap_decomp} decomposes this gap into bucket identification and conditional recovery.
The ground-truth bucket is included in the top-\(S\) buckets selected by GCB for \(57.0\%\) of all queries, but only \(42.4\%\) of baseline errors.
When the baseline is wrong and the ground-truth bucket is missed, GCB cannot promote the target to Top-1, whereas the oracle still reaches \(R@1=71.8\%\).
This shows that a large part of the unrecovered oracle headroom comes from cases where the local logits do not provide sufficiently consistent evidence for the correct sentence bucket.

When the baseline is wrong and the correct bucket is selected, GCB promotes \(39.5\%\) of queries to Top-1.
Hard pruning over the same selected buckets reaches \(R@1=72.7\%\), while the oracle reaches \(R@1=87.2\%\).
Thus, even after bucket identification succeeds, the soft full-pool correction remains conservative: distractor buckets are not removed, the additive Bias may be insufficient to overcome large cross-bucket margins, and within-bucket ordering is unchanged.
This is by design, because GCB is intended as an auditable score-space intervention rather than a hard-pruning or oracle-like procedure.

\begin{table}[!ht]
\centering
\small
\setlength{\tabcolsep}{4pt}
\renewcommand{\arraystretch}{1}
\captionsetup{justification=justified,singlelinecheck=false}
\caption{\textbf{Oracle--GCB headroom decomposition on Gwilliams zero-shot Dense-TCNN.}
Bucket hit denotes whether the ground-truth sentence bucket is included in the top-\(S\) buckets selected by GCB (\(S=3\)).
Conditional metrics are computed on the specified subset.
All retrieval rates are reported as percentages.}
\label{tab:oracle_gap_decomp}

\resizebox{\linewidth}{!}{
\begin{tabular}{lrrrrrr}
\toprule
\textbf{Subset} & \textbf{\# queries} & \textbf{Bucket hit (\%)} & \textbf{Base R@1 (\%)} & \textbf{GCB R@1 (\%)} & \textbf{Hard-prune R@1 (\%)} & \textbf{Oracle R@1 (\%)} \\
\midrule
All queries
& 71,736 & 57.0 & 44.1 & 52.1 & 50.5 & 87.9 \\
Bucket hit
& 40,894 & 100.0 & 58.5 & 74.7 & 88.7 & 94.7 \\
Bucket miss
& 30,842 & 0.0 & 25.1 & 23.6 & 0.0 & 78.9 \\
Baseline errors
& 40,073 & 42.4 & 0.0 & 16.8 & 30.8 & 78.3 \\
Baseline errors, bucket hit
& 16,980 & 100.0 & 0.0 & 39.5 & 72.7 & 87.2 \\
Baseline errors, bucket miss
& 23,093 & 0.0 & 0.0 & 0.0 & 0.0 & 71.8 \\
\bottomrule
\end{tabular}
}
\end{table}

Table~\ref{tab:oracle_gap_rank_bucket} shows the same mechanism from the perspective of baseline rank.
GCB corrects \(38.4\%\) of rank 2--5 errors, \(11.2\%\) of rank 6--10 errors, and only \(0.7\%\) of rank \(>10\) errors.
The ground-truth bucket hit rate also decreases as the baseline rank worsens, from \(56.0\%\) for rank 2--5 to \(30.2\%\) for rank \(>10\).
Thus, poorer local ranking is accompanied by poorer bucket identifiability.
The remaining oracle gap is therefore governed by the quality and margin of the frozen local logits: contextual aggregation can amplify locally competitive evidence, but it cannot replace local stimulus-locked evidence when the correct target or bucket is not supported.

\begin{table}[!ht]
\centering
\small
\setlength{\tabcolsep}{5pt}
\renewcommand{\arraystretch}{1.08}
\captionsetup{justification=justified,singlelinecheck=false}
\caption{\textbf{Top-1 correction by baseline rank bucket on Gwilliams zero-shot Dense-TCNN.}
For baseline errors, we report how often GCB promotes the target to Top-1.
Bucket hit is the fraction of queries whose ground-truth sentence bucket is included in the top-\(S\) buckets selected by GCB.
All rates are reported as percentages.}
\label{tab:oracle_gap_rank_bucket}
\begin{tabular}{lrrrrr}
\toprule
\textbf{Baseline rank} & \textbf{\# queries} & \textbf{Corrected} & \textbf{Correction rate (\%)} & \textbf{Bucket hit (\%)} & \textbf{Oracle R@1 (\%)} \\
\midrule
2--5
& 15,487 & 5,952 & 38.4 & 56.0 & 98.1 \\
6--10
& 5,656 & 634 & 11.2 & 46.1 & 93.7 \\
\(>10\)
& 18,930 & 128 & 0.7 & 30.2 & 57.6 \\
\bottomrule
\end{tabular}
\end{table}

Together, these analyses show that the unrecovered oracle headroom comes from three sources: bucket-identification failures, conservative soft correction under full-pool competition, and residual within-bucket ranking limits.
This is consistent with the role of GCB as a local-evidence-dependent contextual aggregation intervention.

\subsection{Effect of the GCB gain parameter}
\label{app:gamma_sweep}

We characterise the effect of the GCB gain parameter
\(\gamma_{\mathrm{gcb}}\) on the Gwilliams validation split using a single Dense-TCNN checkpoint.
The sweep keeps the local logits, candidate pool, sentence buckets, and all other GCB hyperparameters fixed, and varies only the additive gain applied to the selected sentence buckets.
This isolates how the strength of the score-space correction changes the resulting retrieval decisions.

Table~\ref{tab:gamma_sweep} shows three regimes.
First, the neighbourhood around the reported setting is stable.
For \(\gamma_{\mathrm{gcb}}\in\{0.6,0.7,0.8\}\), GCB gives a positive validation \(\Delta\)R@1 of \(+8.3\) to \(+10.3\) pp, with bad-to-good corrections exceeding good-to-bad regressions by more than \(16{:}1\).
The fixed setting used in the main experiments, \(\gamma_{\mathrm{gcb}}=0.7\), sits in this stable region: it corrects 3,605 baseline errors, demotes 213 baseline hits, changes 14.6\% of Top-1 decisions, and demotes 1.1\% of baseline-correct queries.

Second, neighbouring gains show the local correction--rewrite trade-off.
Reducing the gain to \(\gamma_{\mathrm{gcb}}=0.6\) makes the correction more conservative, with fewer Top-1 rewrites but a smaller \(\Delta\)R@1.
Increasing it to \(\gamma_{\mathrm{gcb}}=0.8\) gives a slightly larger \(\Delta\)R@1, while also changing more local Top-1 decisions.
Thus, the reported setting represents a conservative point within a stable positive-gain neighbourhood.

Third, larger gains move GCB into a more aggressive contextual reweighting regime.
At \(\gamma_{\mathrm{gcb}}=2.2\), \(\Delta\)R@1 rises to \(+0.153\), but good-to-bad regressions increase to 1,050, the correction-to-regression ratio drops from \(16.9{:}1\) to \(6.3{:}1\), and 35.0\% of Top-1 decisions are rewritten.
This indicates that stronger gains recover more validation performance by allowing sentence-bucket context to override a larger fraction of local window-level decisions.
We therefore use \(\gamma_{\mathrm{gcb}}=0.7\) for the reported GCB configuration: it provides substantial contextual correction while keeping the intervention close to a soft full-pool reweighting of the frozen local logits.

\begin{table}[!ht]
\centering
\small
\setlength{\tabcolsep}{5pt}
\renewcommand{\arraystretch}{1.12}
\captionsetup{justification=justified,singlelinecheck=false}
\caption{\textbf{Validation characterisation of the GCB gain parameter on Gwilliams Dense-TCNN.}
The sweep uses a single Dense-TCNN checkpoint on the Gwilliams validation split:
36,358 query windows, \(N_{\mathrm{cand}}=742\) audio candidates, and 2,891 query sentence groups.
All rows use the same frozen local logits, candidate pool, sentence buckets, and GCB rule; only the additive gain is varied.
\(\Delta\)R@1 is measured relative to the local validation baseline R@1 of 52.6\%.
The bold row is the fixed setting used in the reported GCB configuration.
Flip rates are reported as percentages; \(\Delta\)R@1 is reported in percentage points.}
\label{tab:gamma_sweep}
\begin{tabular}{lrrrrrr}
\toprule
\(\boldsymbol{\gamma_{\mathrm{gcb}}}\) &
\(\boldsymbol{\Delta}\)\textbf{R@1 (pp)} &
\textbf{Bad$\rightarrow$good} &
\textbf{Good$\rightarrow$bad} &
\textbf{B2G/G2B} &
\textbf{G2B/base-hit (\%)} &
\textbf{Top-1 changed (\%)} \\
\midrule
0.0 & +0.0  & 0     & 0     & --   & 0.0 & 0.0 \\
0.4 & +5.9  & 2,260 & 112   & 20.2 & 0.6 & 8.8 \\
0.6 & +8.3  & 3,186 & 184   & 17.3 & 1.0 & 12.7 \\
\textbf{0.7} & \textbf{+9.3} & \textbf{3,605} & \textbf{213} & \textbf{16.9} & \textbf{1.1} & \textbf{14.6} \\
0.8 & +10.3 & 3,974 & 244   & 16.3 & 1.3 & 16.3 \\
1.1 & +12.6 & 4,952 & 358   & 13.8 & 1.9 & 21.3 \\
1.4 & +14.1 & 5,644 & 529   & 10.7 & 2.8 & 25.6 \\
2.2 & +15.3 & 6,606 & 1,050 & 6.3  & 5.5 & 35.0 \\
\bottomrule
\end{tabular}
\end{table}

\subsection{Session-isolated stress test under expanded candidate scale}
\label{app:slr}

We test the Base--GCB contrast on a harder Gwilliams evaluation regime in which recording-session transfer and candidate-pool scale are stressed simultaneously.
Split construction is recording-level: each \((\texttt{subject\_id}, \texttt{session\_id})\) pair is assigned to exactly one of \(\{\texttt{train}, \texttt{valid}, \texttt{test}\}\), targeting a \(0.7/0.1/0.2\) recording ratio via largest-remainder rounding with \(\pm 1\) recording tolerance.
To make session transfer rather than novel stimulus identity the primary difficulty, every validation and test subject appears in training at least once, and test windows are filtered to stimulus content covered by training.
For \(k{=}5\) folds, we impose a subject-rotation constraint: subjects with at least two recordings are partitioned into \(k\) groups, and fold \(i\) assigns at most one recording per subject from group \(i\) to the test split while retaining at least one recording from the same subject in training.
Each fold uses seed \((\texttt{SEED0}+i)\), with \(\texttt{SEED0}=42\); reported metrics are averaged over five folds and three model seeds.

This regime stresses two attribution-relevant factors jointly.
First, local evidence must generalise across recording sessions.
Second, the candidate pool expands to \(N{=}7{,}011\), which is \(4.8{\times}\) larger than the Gwilliams zero-shot pool (\(N{=}1{,}464\)), substantially increasing cross-bucket competition for the contextual source.
We treat this setting as a boundary-condition audit: it does not replace the zero-shot attribution endpoint, but probes how far the Base--GCB contrast persists when local ranking is weakened and candidate-scale competition is increased.

Table~\ref{tab:gcb_main_sessiso} reports the results.
The local baseline degrades substantially: Dense-TCNN R@10 drops from \(73.6\%\) at \(N{=}1{,}464\) to \(37.4\%\) at \(N{=}7{,}011\).
The Base--GCB contrast contracts accordingly, from \(+7.7\) R@1 points in Gwilliams zero-shot to \(+0.59\) points here.
This indicates that contextual aggregation remains measurable, but its recoverable contribution shrinks sharply once local logits are weakened and correct targets are less often competitive under the expanded candidate pool.
The resulting ceiling is imposed by the frozen local evidence rather than by the contextual score correction alone.

\providecommand{\sd}[2]{\ensuremath{#1{\scriptstyle\pm}#2}}
\providecommand{\bsd}[2]{\ensuremath{\mathbf{#1}{\scriptstyle\pm}\mathbf{#2}}}
\providecommand{\bn}[1]{\ensuremath{\mathbf{#1}}}

\begin{table*}[!ht]
  \centering
  \setlength{\tabcolsep}{2.2pt}
  \renewcommand{\arraystretch}{0.98}
  \scriptsize
  \captionsetup{justification=justified,singlelinecheck=false}
  \caption{\textbf{GCB-over-Base contrast under Gwilliams session-isolated candidate-scale stress.}
This table audits the recoverable contextual source when recording sessions are isolated and the candidate pool expands to \(N{=}7{,}011\).
Each backbone is evaluated before and after applying GCB to the same frozen retrieval logits, so the contrast holds the encoder, candidate pool, and local logits fixed.
Dense denotes Dense-TCNN; Exp-5 is the receptive-field-matched exp-dilated reference of~\citet{defossez2023decoding}; Exp-10 is the deeper exp-dilated reference-depth variant.
Arrows indicate the preferred direction.
R@1, MRR, and R@10 are reported as percentages; \(\Delta\) columns report absolute percentage-point gains.
R@1, MRR, R@10, and \(\Delta\) are mean\(\pm\)std over 15 fold--seed runs (\(5\) folds \(\times\) \(3\) seeds), while MedR is the mean over runs.
The std reflects both session-split variation and model-seed variation.
Bold: the best value within each column.}
  \label{tab:gcb_main_sessiso}

  \resizebox{0.99\textwidth}{!}{%
  \begin{tabular}{@{}lcccccccccc@{}}
    \toprule
    \multirow{2}{*}{\textbf{Backbone}} &
    \multicolumn{4}{c}{\textbf{Base}} &
    \multicolumn{4}{c}{\textbf{+GCB}} &
    \multicolumn{2}{c}{\textbf{\(\Delta\)}} \\
    \cmidrule(lr){2-5}
    \cmidrule(lr){6-9}
    \cmidrule(l){10-11}
    &
    \textbf{R@1 (\%)\(\uparrow\)} & \textbf{MRR (\%)\(\uparrow\)} & \textbf{R@10 (\%)\(\uparrow\)} & \textbf{MedR\(\downarrow\)} &
    \textbf{R@1 (\%)\(\uparrow\)} & \textbf{MRR (\%)\(\uparrow\)} & \textbf{R@10 (\%)\(\uparrow\)} & \textbf{MedR\(\downarrow\)} &
    \textbf{\(\Delta\)R@1 (pp)\(\uparrow\)} & \textbf{\(\Delta\)MRR (pp)\(\uparrow\)} \\
    \midrule

    Dense-TCNN
      & \bsd{17.0}{3.5} & \bsd{24.0}{3.8} & \bsd{37.4}{4.6} & \bn{33.9}
      & \bsd{17.6}{3.5} & \bsd{24.6}{3.9} & \bsd{38.0}{4.6} & \bn{32.7}
      & \bsd{0.59}{0.22} & \bsd{0.64}{0.18} \\

    Exp-dilated ($L=5$)
      & \sd{16.2}{3.0} & \sd{23.0}{3.3} & \sd{36.1}{4.0} & 38.1
      & \sd{16.7}{3.1} & \sd{23.6}{3.4} & \sd{36.8}{4.1} & 37.0
      & \sd{0.52}{0.27} & \sd{0.59}{0.23} \\

    Exp-dilated ($L=10$)
      & \sd{15.1}{2.5} & \sd{21.7}{2.8} & \sd{34.5}{3.6} & 42.9
      & \sd{15.4}{2.6} & \sd{22.1}{2.9} & \sd{35.1}{3.7} & 42.0
      & \sd{0.37}{0.33} & \sd{0.45}{0.27} \\
    \bottomrule
  \end{tabular}%
  }
  \vspace{-1em}
\end{table*}

Table~\ref{tab:sessiso_fold_delta} breaks down the Base--GCB contrast by fold to characterise its dependence on the held-out session partition.
Fold-level averages are positive in all reported entries, but the magnitude varies considerably, particularly for the deeper exp-dilated variant.
This indicates that the recoverable contextual signal under session isolation is sensitive to which recording sessions are withheld.

\begin{table*}[!ht]
  \centering
  \scriptsize
  \setlength{\tabcolsep}{4pt}
  \renewcommand{\arraystretch}{1.05}
  \captionsetup{justification=justified,singlelinecheck=false}
  \caption{\textbf{Fold-level GCB-over-Base gains under Gwilliams session-isolated evaluation.}
Each cell reports mean\(\pm\)std over three model seeds within the fold.
Values are absolute percentage-point differences after applying GCB to the frozen Base logits.
This table summarises fold-level heterogeneity without expanding all 15 fold--seed runs.}
  \label{tab:sessiso_fold_delta}

  \begin{tabular}{@{}c cc cc cc@{}}
    \toprule
    \multirow{2}{*}{\textbf{Fold}} &
    \multicolumn{2}{c}{\textbf{Dense-TCNN}} &
    \multicolumn{2}{c}{\textbf{Exp-dilated ($L=5$)}} &
    \multicolumn{2}{c}{\textbf{Exp-dilated ($L=10$)}} \\
    \cmidrule(lr){2-3}
    \cmidrule(lr){4-5}
    \cmidrule(l){6-7}
    &
    \textbf{\(\Delta\)R@1 (pp)} & \textbf{\(\Delta\)MRR (pp)} &
    \textbf{\(\Delta\)R@1 (pp)} & \textbf{\(\Delta\)MRR (pp)} &
    \textbf{\(\Delta\)R@1 (pp)} & \textbf{\(\Delta\)MRR (pp)} \\
    \midrule
    1 & \sd{0.45}{0.05} & \sd{0.52}{0.04} & \sd{0.26}{0.09} & \sd{0.36}{0.09} & \sd{0.27}{0.28} & \sd{0.36}{0.23} \\
    2 & \sd{0.77}{0.18} & \sd{0.81}{0.14} & \sd{0.74}{0.15} & \sd{0.78}{0.12} & \sd{0.31}{0.23} & \sd{0.43}{0.22} \\
    3 & \sd{0.61}{0.17} & \sd{0.65}{0.13} & \sd{0.72}{0.24} & \sd{0.77}{0.19} & \sd{0.33}{0.33} & \sd{0.43}{0.28} \\
    4 & \sd{0.69}{0.36} & \sd{0.71}{0.30} & \sd{0.66}{0.04} & \sd{0.69}{0.05} & \sd{0.61}{0.51} & \sd{0.64}{0.43} \\
    5 & \sd{0.43}{0.06} & \sd{0.52}{0.07} & \sd{0.22}{0.11} & \sd{0.33}{0.11} & \sd{0.33}{0.37} & \sd{0.42}{0.31} \\
    \bottomrule
  \end{tabular}
\end{table*}

The flip and rank-bucket diagnostics in Tables~\ref{tab:session_flip}--\ref{tab:session_rank_bucket} characterise the residual contextual source more directly.
Bad-to-good Top-1 flips consistently outnumber good-to-bad regressions across all backbones, showing that the small aggregate contrast reflects directional correction rather than symmetric score noise.

\begin{table}[!ht]
\centering
\small
\setlength{\tabcolsep}{5pt}
\renewcommand{\arraystretch}{1.12}
\captionsetup{justification=justified,singlelinecheck=false}
\caption{\textbf{Top-1 flip analysis in Gwilliams session-isolation.}
Results are aggregated over five folds and three seeds.
A bad-to-good flip denotes a query that is not Top-1 under the baseline but becomes Top-1 after GCB.
A good-to-bad flip denotes a query that is Top-1 under the baseline but is demoted after GCB.
Net gain rate is defined as \((\mathrm{bad{\rightarrow}good}-\mathrm{good{\rightarrow}bad})/(\mathrm{bad{\rightarrow}good}+\mathrm{good{\rightarrow}bad})\).}
\label{tab:session_flip}
\begin{tabular}{lrrrr}
\toprule
\textbf{Backbone} & \textbf{Good$\rightarrow$bad} & \textbf{Bad$\rightarrow$good} & \textbf{Net corrections} & \textbf{Net gain rate (\%)} \\
\midrule
Dense-TCNN              & 10,621 & 17,726 & +7,105 & 25.1 \\
Exp-dilated $(L{=}5)$  & 9,992  & 15,451 & +5,459 & 21.5 \\
Exp-dilated $(L{=}10)$ & 9,960  & 13,837 & +3,877 & 16.3 \\
\bottomrule
\end{tabular}
\end{table}

The correction signal is concentrated near the top of the baseline ranking.
Table~\ref{tab:session_rank_bucket} shows that, for Dense-TCNN, \(9.3\%\) of rank 2--5 errors and \(3.8\%\) of rank 6--10 errors are promoted to Top-1, whereas rank \(>10\) errors are almost never recovered.
The expanded candidate pool therefore shifts the regime mainly by moving correct targets beyond this recoverable range, rather than by disrupting the aggregation mechanism itself.

\begin{table}[!ht]
\centering
\small
\setlength{\tabcolsep}{4pt}
\renewcommand{\arraystretch}{1.12}
\captionsetup{justification=justified,singlelinecheck=false}
\caption{\textbf{Top-1 correction rate by baseline rank bucket in Gwilliams session-isolation.}
For queries where the correct answer is not ranked Top-1 by the baseline, we report how often GCB promotes it to Top-1.
Format: corrected / total (rate).}
\label{tab:session_rank_bucket}
\begin{tabular}{lccc}
\toprule
\textbf{Backbone} & \textbf{Rank 2--5} & \textbf{Rank 6--10} & \textbf{Rank \(>10\)} \\
\midrule
Dense-TCNN
& 13,653 / 146,180 (9.3\%)
& 2,692 / 70,508 (3.8\%)
& 1,381 / 629,830 (0.2\%) \\
Exp-dilated $(L{=}5)$
& 11,771 / 139,584 (8.4\%)
& 2,332 / 70,202 (3.3\%)
& 1,348 / 671,486 (0.2\%) \\
Exp-dilated $(L{=}10)$
& 10,472 / 134,983 (7.8\%)
& 2,246 / 70,080 (3.2\%)
& 1,119 / 688,308 (0.2\%) \\
\bottomrule
\end{tabular}
\end{table}

This pattern defines the boundary condition of the contextual source.
Cross-window aggregation can amplify sentence-consistent local evidence when the frozen logits already place the correct target within competitive range; it cannot substitute for missing stimulus-locked signal or overcome arbitrary full-pool competition.
The contraction of the Base--GCB contrast under session isolation therefore locates the ceiling of the contextual source in the frozen local evidence rather than in the contextual score correction.

\subsection{Length-dependent gain analysis}
\label{app:length_dependence}

\begin{figure}[!ht]
    \centering
    \includegraphics[width=0.7\linewidth]{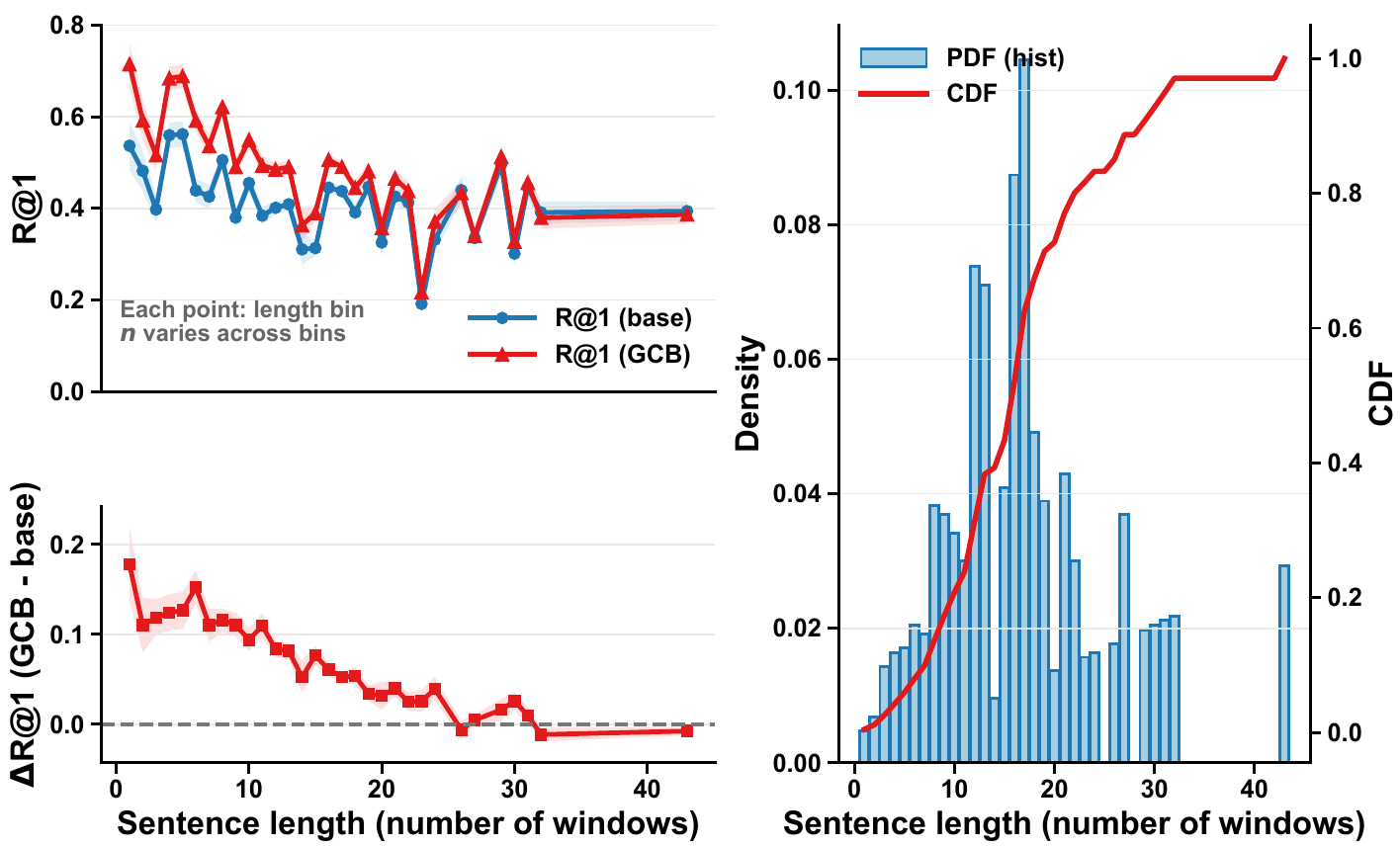}
    \caption{\textbf{Contextual aggregation effect by sentence length on Gwilliams zero-shot.}
    Left: binned R@1 for the local baseline and the GCB-corrected retrieval scores.
    Bottom-left: \(\Delta\)R@1 per bin induced by GCB.
    Right: sentence-length distribution.
    The analysis characterises the operating regime of the contextual effect rather than estimating a causal effect of sentence length, since sentence length is coupled with bucket size, sentence heterogeneity, and within-bucket difficulty.}
    \label{fig:R@1_vs_length}
\end{figure}

Fig.~\ref{fig:R@1_vs_length} reports how the Base--GCB contrast varies with sentence length on Gwilliams zero-shot retrieval.
Sentence length is measured by the number of evaluation windows in the sentence.
The analysis should be interpreted as an operating-regime characterisation rather than as a causal estimate of sentence length alone, because sentence length is coupled with the number of query windows, candidate-bucket size, within-bucket difficulty, sentence heterogeneity, and the number of sentence samples per bin.

The observed pattern is consistent with the mechanism in Sec.~\ref{subsec:mtd_context}.
GCB pools sentence-consistent evidence under a shared sentence context, so it benefits most when neighbouring windows provide compatible support for the same candidate buckets.
For short-to-medium sentences, this consistency is often strong enough to support a positive Base--GCB contrast.
For longer sentences, the available evidence can become more heterogeneous: the sentence spans more linguistic content, bucket size increases, and within-bucket ranking becomes harder.
The reduced gain in the longest bins should therefore be read as a boundary of the operating regime, not as evidence that sentence length alone causally determines the contextual effect.

\subsection{Within-MEG evidence attenuation}
\label{app:evidence_attenuation}

This diagnostic tests how the contextual gain behaves as target-specific local evidence is progressively weakened while the dataset, encoder, candidate pool, sentence grouping, and GCB hyperparameters are held fixed.
It is designed as an evidence-attenuation diagnostic for source attribution.

\paragraph{Interpretation and robustness logic.}
The real-MEG mixture provides a controlled way to attenuate target-specific evidence while preserving realistic neural statistics.
The donor window \(\tilde{x}_w\) is sampled from a different sentence key, so it may introduce distractor neural structure but should not systematically support the current target candidate or its sentence bucket.
We also test three complementary surrogate families that degrade local evidence through different mechanisms: logit-space permutation, phase randomisation, and covariance-matched Gaussian noise.
Across these perturbation families, the diagnostic asks whether the Base--GCB contrast vanishes as achieved local retrieval approaches chance.

\paragraph{Real-MEG stimulus-mismatched attenuation.}
For each query window \(x_w\), we sample a stimulus-mismatched real MEG window \(\tilde{x}_w\) from the same evaluation set, preferring the same subject when possible, and construct
\begin{equation}
x_w^{(\alpha)}
=
\sqrt{\alpha}\,x_w
+
\sqrt{1-\alpha}\,\tilde{x}_w ,
\qquad
\alpha\in\{0.75,0.50,0.25,0.10,0.00\}.
\label{eq:attenuation_mix}
\end{equation}
Here \(\alpha{=}1\) is the original query, whereas \(\alpha{=}0\) contains only stimulus-mismatched MEG with no target-specific evidence for the current candidate.

\paragraph{Additional surrogate families.}
We also test three complementary perturbations to avoid relying on the statistical assumptions of real-MEG mixing.
First, a logit-space permutation diagnostic weakens candidate-specific evidence after encoding by mixing clean logits with row-wise candidate-permuted logits; this bypasses neural-signal statistics entirely.
Second, a phase-randomised MEG surrogate preserves channel-wise spectral magnitude while disrupting time-locked structure.
Third, a spatial-covariance-matched Gaussian surrogate uses an empirical sensor covariance estimated from real MEG windows.
These variants degrade local evidence at different levels of the pipeline, so convergence across them supports the evidence-dependence conclusion rather than a specific noise-model assumption.

\paragraph{Perturbation implementation details.}
All neural-space perturbations are applied after the shared preprocessing pipeline and before the frozen neural encoder.
For real-MEG mixing, donor indices are built once with a fixed random seed.
For each query window, donors are sampled with replacement. We first sample from windows belonging to the same subject and require a different sentence key; if no such donor is found after repeated attempts, we sample from all evaluation windows with a different sentence key.
No additional rescaling is applied for real-MEG mixing beyond the shared preprocessing pipeline.

For logit permutation, we perturb the already-computed clean local logits rather than the neural input.
For each query row, candidate scores are independently permuted, and both clean and permuted rows are standardised before mixing:
\begin{equation}
\tilde{L}^{(\alpha)}
=
\mu_L+\sigma_L
\left(
\sqrt{\alpha}\,Z_L+\sqrt{1-\alpha}\,Z_{\mathrm{perm}}
\right).
\label{eq:logit_permutation_attenuation}
\end{equation}
where \(Z_L\) is the row-standardised clean logit vector, \(Z_{\mathrm{perm}}\) is the row-standardised permuted logit vector, and \(\mu_L,\sigma_L\) are the mean and standard deviation of the clean row.
This preserves the original row scale while progressively weakening candidate-specific evidence.

For phase randomisation, we apply a fast Fourier transform (FFT) along the time dimension of each channel, keep the amplitude spectrum fixed, replace the phase values with independent random phases except at the zero-frequency component and the Nyquist-frequency component, and then invert the transform to obtain a real-valued surrogate window.
The phase-randomised window is then mixed with the original window using Eq.~\eqref{eq:attenuation_mix}.

For covariance-matched Gaussian perturbation, we estimate a spatial covariance matrix from 2,048 randomly sampled preprocessed MEG windows using a fixed seed.
The covariance is estimated separately for each channel count after subtracting the per-channel temporal mean, and a small ridge term is added before Cholesky factorisation to generate covariance-matched Gaussian samples.
For each query batch, we sample zero-mean Gaussian noise with this spatial covariance, match its root-mean-square (RMS) amplitude to the original query window, and then apply Eq.~\eqref{eq:attenuation_mix}.

\begin{figure}[!ht]
  \centering
  \includegraphics[width=0.7\linewidth]{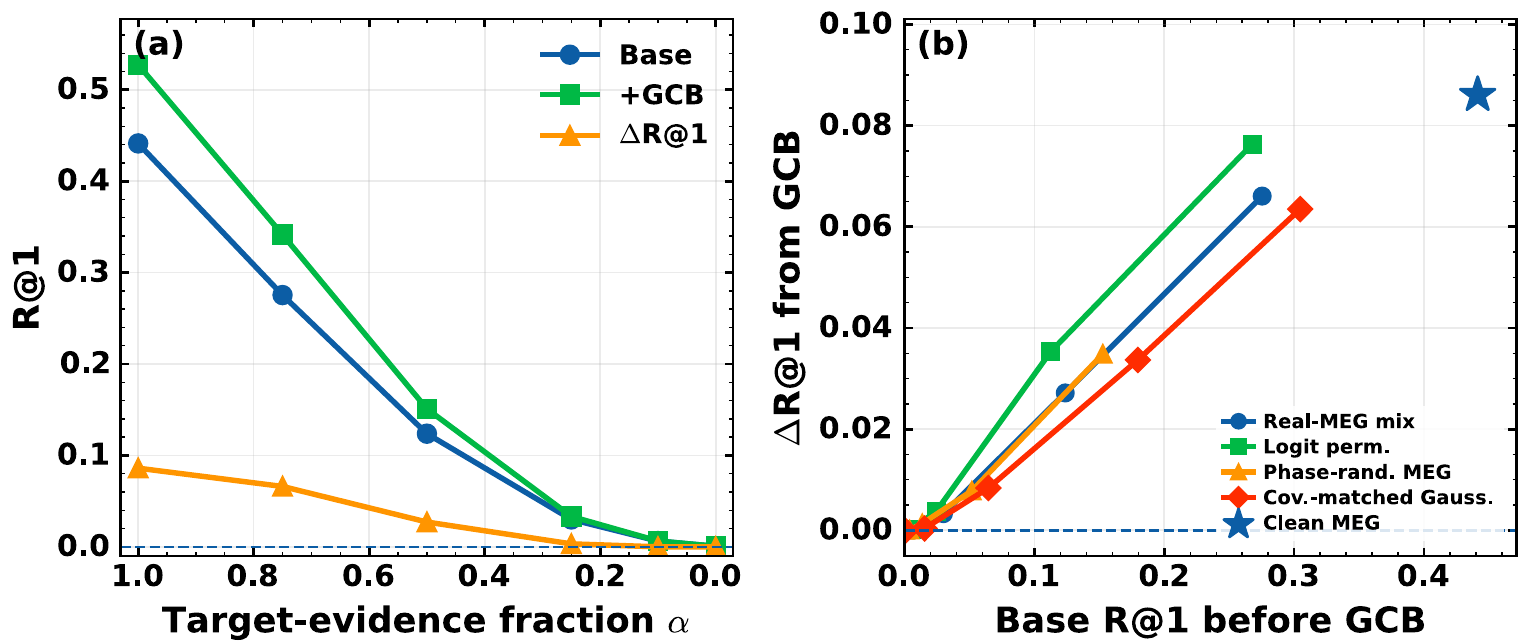}
  \caption{\textbf{Contextual gain tracks available local evidence under within-Gwilliams attenuation.}
  (a) Real-MEG stimulus-mismatched attenuation. Each attenuated query mixes the original MEG window with a stimulus-mismatched real MEG window from the same evaluation set. As the target-evidence fraction \(\alpha\) decreases, base retrieval, +GCB retrieval, and the contextual gain all collapse.
  (b) Evidence attenuation across surrogate families. Each point is one attenuation condition. The x-axis shows base retrieval R@1 before GCB, and the y-axis shows query-weighted \(\Delta\)R@1 after GCB. Across perturbation families, contextual gains vanish as local retrieval approaches chance.}
  \label{fig:app_evidence_attenuation}
\end{figure}

\paragraph{Results.}
Fig.~\ref{fig:app_evidence_attenuation} summarises the attenuation suite.
(a) shows the primary real-MEG attenuation result: as \(\alpha\) decreases, both base retrieval and +GCB retrieval decline, and the contextual gain collapses with the available local evidence.
(b) shows the same relationship across perturbation families by plotting \(\Delta\)R@1 against base R@1 before GCB.
The same value of \(\alpha\) is not directly comparable across perturbation types, so base R@1 is the more informative horizontal axis.
Across real-MEG mixing, logit permutation, phase-randomised MEG, and covariance-matched Gaussian surrogates, contextual gains vanish as local retrieval approaches chance.

Table~\ref{tab:evidence_attenuation_realmix} reports the primary real-MEG attenuation result numerically.
As target-specific evidence is reduced, baseline R@1 decreases and the GCB gain tracks the available local evidence: query-weighted \(\Delta\)R@1 is \(+8.6\) pp on clean MEG, \(+6.6\) pp at \(\alpha{=}0.75\), \(+2.7\) pp at \(\alpha{=}0.50\), \(+0.3\) pp at \(\alpha{=}0.25\), and approximately zero at \(\alpha{\le}0.10\).
This supports the source-attribution interpretation: GCB separates contextual aggregation from encoder learning, but the contextual source becomes measurable only when local stimulus-locked evidence is present.

\begin{table}[!ht]
\centering
\small
\setlength{\tabcolsep}{6pt}
\renewcommand{\arraystretch}{1.12}
\caption{\textbf{Within-Gwilliams real-MEG evidence attenuation.}
Each attenuated query mixes the original MEG window with a stimulus-mismatched real MEG window from the same evaluation set.
The encoder, candidate pool, sentence grouping, and GCB hyperparameters are fixed.
Values are query-weighted over \(Q{=}71{,}736\) windows from a single controlled diagnostic run.
All retrieval metrics are reported as percentages; deltas denote absolute percentage-point differences computed before rounding.}
\label{tab:evidence_attenuation_realmix}
\begin{tabular}{lcccc}
\toprule
\textbf{Condition} &
\textbf{Base R@1 (\%)} &
\textbf{+GCB R@1 (\%)} &
\(\boldsymbol{\Delta}\)\textbf{R@1 (pp)} &
\(\boldsymbol{\Delta}\)\textbf{MRR (pp)} \\
\midrule
Clean MEG                         & 44.1 & 52.7 & +8.6 & +7.2 \\
Real-MEG mix \(\alpha{=}0.75\)    & 27.5 & 34.2 & +6.6 & +6.1 \\
Real-MEG mix \(\alpha{=}0.50\)    & 12.4 & 15.1 & +2.7 & +2.8 \\
Real-MEG mix \(\alpha{=}0.25\)    & 3.0  & 3.3  & +0.3 & +0.4 \\
Real-MEG mix \(\alpha{=}0.10\)    & 0.6  & 0.6  & +0.0 & +0.0 \\
Real-MEG mix \(\alpha{=}0.00\)    & 0.1  & 0.1  & +0.0 & +0.0 \\
\bottomrule
\end{tabular}
\end{table}

\paragraph{Sentence-level bootstrap.}
Table~\ref{tab:evidence_attenuation_bootstrap_realmix} reports sentence-level paired bootstrap estimates for the same real-MEG attenuation diagnostic.
The numerical deltas differ from Table~\ref{tab:evidence_attenuation_realmix}, since sentence-level averaging weights short and long sentences differently from query-weighted metrics.
The qualitative pattern is unchanged: sizeable GCB gains appear only when local retrieval remains informative, and the effect collapses as target-specific evidence is removed.
Stars in Table~\ref{tab:evidence_attenuation_bootstrap_realmix} indicate paired-bootstrap significance, but significance is not the relevant operating-regime criterion at very low evidence levels.
For \(\alpha\le 0.10\), the statistically detectable differences are only about \(0.3\) percentage points in R@1/MRR or smaller, and are therefore not substantively meaningful for retrieval.

\begin{table}[!ht]
\centering
\small
\setlength{\tabcolsep}{6pt}
\renewcommand{\arraystretch}{1.12}
\caption{\textbf{Sentence-level paired bootstrap for real-MEG evidence attenuation.}
\(\Delta\) denotes +GCB minus the local baseline, computed with paired sentence-cluster bootstrap over 123 sentence buckets.
Intervals are 95\% bootstrap confidence intervals.
Stars indicate paired-bootstrap significance (\({}^{*}p{<}0.001\)).
All deltas and confidence intervals are reported as absolute percentage-point differences.}
\label{tab:evidence_attenuation_bootstrap_realmix}
\begin{tabular}{lccc}
\toprule
\textbf{Condition} & \textbf{Metric} & \(\boldsymbol{\Delta}\textbf{ (pp)}\) & \textbf{95\% CI (pp)} \\
\midrule
Clean MEG & R@1 & \(+10.78^{*}\) & [9.99, 11.59] \\
Clean MEG & MRR & \(+8.94^{*}\)  & [8.31, 9.57] \\
\midrule
Real-MEG mix \(\alpha{=}0.75\) & R@1 & \(+9.34^{*}\) & [8.43, 10.30] \\
Real-MEG mix \(\alpha{=}0.75\) & MRR  & \(+8.49^{*}\) & [7.73, 9.28] \\
\midrule
Real-MEG mix \(\alpha{=}0.50\) & R@1 & \(+5.06^{*}\) & [4.27, 5.93] \\
Real-MEG mix \(\alpha{=}0.50\) & MRR  & \(+5.07^{*}\) & [4.35, 5.84] \\
\midrule
Real-MEG mix \(\alpha{=}0.25\) & R@1 & \(+1.06^{*}\) & [0.74, 1.44] \\
Real-MEG mix \(\alpha{=}0.25\) & MRR  & \(+1.28^{*}\) & [0.93, 1.68] \\
\midrule
Real-MEG mix \(\alpha{=}0.10\) & R@1 & \(+0.21^{*}\) & [0.08, 0.38] \\
Real-MEG mix \(\alpha{=}0.10\) & MRR  & \(+0.31^{*}\) & [0.17, 0.48] \\
\midrule
Real-MEG mix \(\alpha{=}0.00\) & R@1 & \(+0.02\) & [0.00, 0.05] \\
Real-MEG mix \(\alpha{=}0.00\) & MRR  & \(+0.02\) & [0.00, 0.05] \\
\bottomrule
\end{tabular}
\end{table}

\paragraph{Other attenuation families.}
Table~\ref{tab:evidence_attenuation_all} reports the additional surrogate families.
Although the same value of \(\alpha\) does not produce identical degradation across perturbation types, all families show the same qualitative relationship: GCB gain is large when base retrieval remains informative and collapses as local evidence approaches chance.
This supports interpreting GCB as amplifying sentence-consistent local evidence rather than exploiting sentence grouping by itself.
\begin{table}[!ht]
\centering
\small
\setlength{\tabcolsep}{7pt}
\renewcommand{\arraystretch}{1.10}
\caption{\textbf{Evidence attenuation across surrogate families.}
Values are query-weighted.
The same \(\alpha\) is not directly comparable across perturbation types; the relevant pattern is that \(\Delta\)R@1 decreases as baseline R@1 approaches chance.
Retrieval rates are reported as percentages; deltas denote absolute percentage-point differences.}
\label{tab:evidence_attenuation_all}
\begin{tabular}{lccc}
\toprule
\textbf{Evidence level} &
\textbf{Baseline R@1 (\%)} &
\textbf{GCB-corrected R@1 (\%)} &
\(\boldsymbol{\Delta}\)\textbf{R@1 (pp)} \\
\midrule
\multicolumn{4}{l}{\emph{Covariance-matched Gaussian perturbation}} \\
\(\alpha{=}0.75\) & 30.5 & 36.8 & +6.4 \\
\(\alpha{=}0.50\) & 18.0 & 21.3 & +3.4 \\
\(\alpha{=}0.25\) & 6.5  & 7.3  & +0.8 \\
\(\alpha{=}0.10\) & 1.6  & 1.6  & +0.0 \\
\(\alpha{=}0.00\) & 0.1  & 0.1  & +0.0 \\
\midrule
\multicolumn{4}{l}{\emph{Logit-space permutation perturbation}} \\
\(\alpha{=}0.75\) & 26.8 & 34.4 & +7.6 \\
\(\alpha{=}0.50\) & 11.3 & 14.8 & +3.5 \\
\(\alpha{=}0.25\) & 2.4  & 2.8  & +0.4 \\
\(\alpha{=}0.10\) & 0.6  & 0.6  & +0.0 \\
\(\alpha{=}0.00\) & 0.1  & 0.1  & +0.0 \\
\midrule
\multicolumn{4}{l}{\emph{Phase-randomised MEG perturbation}} \\
\(\alpha{=}0.75\) & 15.3 & 18.8 & +3.5 \\
\(\alpha{=}0.50\) & 5.2  & 6.0  & +0.8 \\
\(\alpha{=}0.25\) & 1.4  & 1.5  & +0.1 \\
\(\alpha{=}0.10\) & 0.4  & 0.4  & +0.0 \\
\(\alpha{=}0.00\) & 0.1  & 0.1  & +0.0 \\
\bottomrule
\end{tabular}
\end{table}

\subsection{Brennan EEG: an out-of-distribution evidence-limited stress test}
\label{app:brennan_pool_sweep}

The within-Gwilliams attenuation diagnostic in Sec.~\ref{app:evidence_attenuation} provides the controlled test: local evidence is weakened while the dataset, encoder, candidate pool, sentence grouping, and GCB hyperparameters are fixed.
Brennan EEG evaluates the same aggregation rule in a realistic out-of-distribution evidence-limited regime.
Since it differs from the MEG experiments in modality, participants, and stimulus material, we treat it as a boundary-condition test for score-space aggregation under weak local retrieval evidence.

\begin{table}[!ht]
\centering
\footnotesize
\setlength{\tabcolsep}{3pt}
\renewcommand{\arraystretch}{1.15}
\caption{\textbf{EEG retrieval on Brennan (\(N{=}388\)).}
Performance is close to chance under fixed-length closed-set retrieval, and GCB yields no reliable gains, consistent with unreliable bucket support in an evidence-limited regime.
Parentheses report the multiple over the corresponding random-chance baseline.
All retrieval metrics are reported as percentages.}
\label{tab:eeg_results}
\begin{tabular}{lcccc}
\toprule
\textbf{Backbone} &
\textbf{R@1 (\%)} &
\textbf{R@5 (\%)} &
\textbf{R@10 (\%)} &
\textbf{MRR (\%)} \\
\midrule
Random chance      & 0.26 & 1.29 & 2.58 & 1.69 \\
\midrule
Exp-dilated        & 0.38 (\(1.5{\times}\)) & 1.72 (\(1.3{\times}\)) & 3.47 (\(1.3{\times}\)) & 2.04 (\(1.2{\times}\)) \\
Exp-dilated + GCB  & 0.31 (\(1.2{\times}\)) & 1.69 (\(1.3{\times}\)) & 3.39 (\(1.3{\times}\)) & 1.98 (\(1.2{\times}\)) \\
Dense-TCNN         & 0.72 (\(2.8{\times}\)) & 3.59 (\(2.8{\times}\)) & 7.21 (\(2.8{\times}\)) & 3.88 (\(2.3{\times}\)) \\
Dense-TCNN + GCB   & 0.21 (\(0.8{\times}\)) & 1.18 (\(0.9{\times}\)) & 2.37 (\(0.9{\times}\)) & 1.57 (\(0.9{\times}\)) \\
\bottomrule
\end{tabular}
\end{table}

Table~\ref{tab:eeg_results} reports retrieval on the full Brennan EEG candidate pool
(\(N{=}388\); random \(R@1\approx 1/N=0.0026\)).
Both backbones operate in a near-chance regime: Dense-TCNN is slightly above random, but its absolute retrieval accuracy remains very low compared with the MEG settings.
We therefore treat Brennan as a low-evidence endpoint for this retrieval protocol.
This is consistent with the expected difficulty of EEG relative to MEG in recovering fine-grained time-locked linguistic information, while also reflecting dataset, participant, and stimulus differences.

The negative GCB deltas are consistent with this evidence-limited regime.
GCB adds a bucket-wise bias to the sentence buckets that receive the strongest support from the frozen local logits.
When local logits contain coherent stimulus-locked evidence, this bias can correct cross-sentence near misses, as shown by the Gwilliams flip analysis.
When local logits are close to chance, the top-supported buckets are often spurious, so the same correction can amplify incorrect sentence hypotheses and demote occasional correct local hits.

This failure mode is consistent with the component ablations in Table~\ref{tab:gcb_ablation}.
GCB-single already falls below the local baseline, and removing the high-evidence gate degrades performance further, showing that aggregation requires selective and reliable bucket support.
Brennan EEG represents a stronger version of the same condition: the local logits provide too little stable evidence for cross-window pooling to recover a consistent true sentence bucket.
This marks a boundary condition of the intervention: contextual aggregation amplifies sentence-consistent local evidence when such evidence is present.

\begin{figure}[!ht]
  \centering
  \includegraphics[width=1\linewidth]{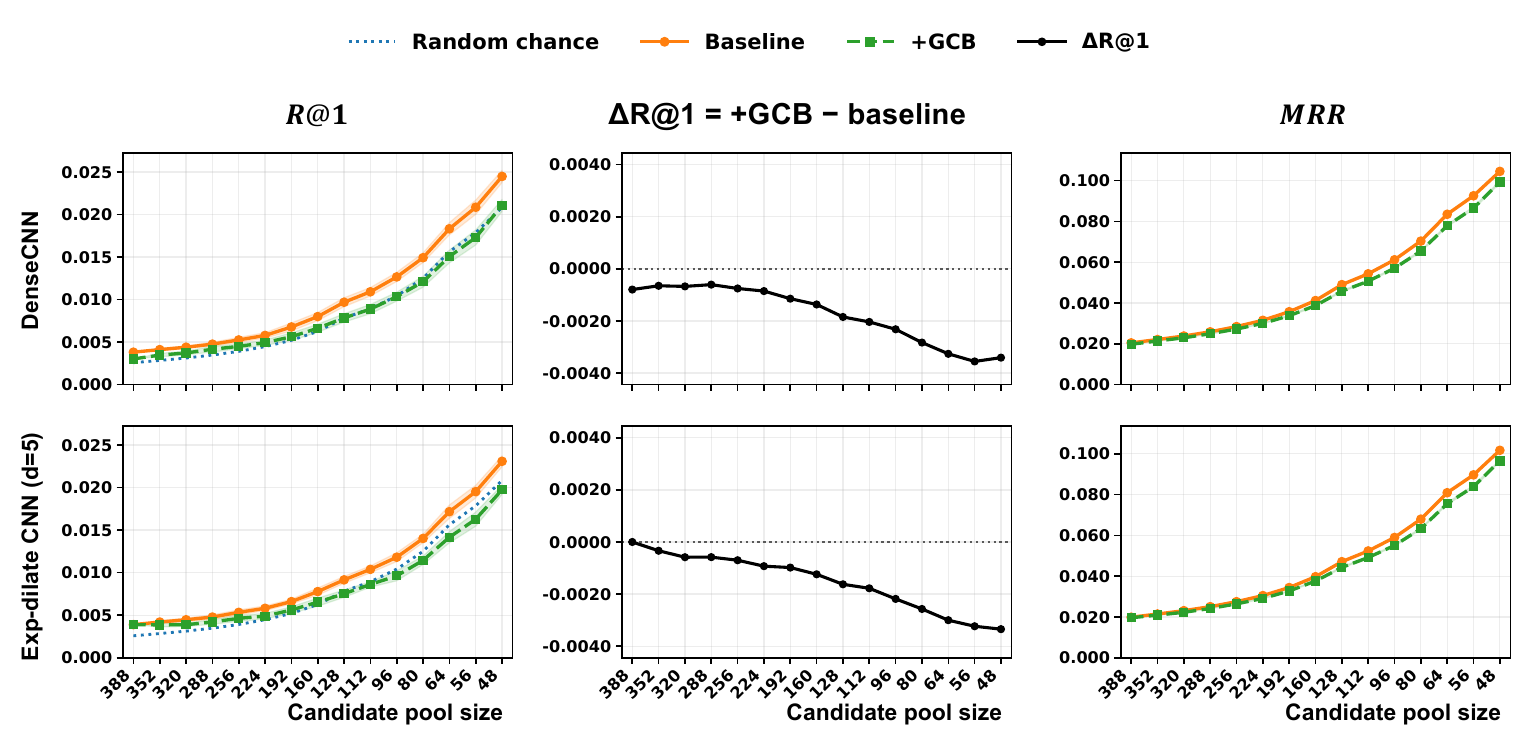}
  \caption{\textbf{Brennan EEG retrieval performance across candidate-pool sizes.}
  Absolute R@1 and MRR improve as the candidate pool shrinks, but \(\Delta\mathrm{R@1} = \mathrm{GCB} - \mathrm{baseline}\) remains negative across pool sizes for both backbones.
  This supports interpreting Brennan as an out-of-distribution evidence-limited regime: reducing pool difficulty raises absolute retrieval scores, but does not create reliable sentence-bucket support for contextual aggregation.}
  \label{fig:app_brennan_pool_sweep}
\end{figure}

We further evaluate Brennan EEG retrieval across candidate-pool sizes to separate the effect of pool difficulty from the availability of decodable neural evidence.
As the candidate pool shrinks from \(N{=}388\) to smaller subsets, absolute R@1 and MRR improve for both Dense-TCNN and Exp-dilated backbones.
However, the GCB improvement remains negative across pool sizes (Fig.~\ref{fig:app_brennan_pool_sweep}).
This pattern suggests that the absence of GCB gain is not explained only by the large candidate pool.
Instead, even when the closed-set task is made easier, the local logits still do not provide reliable sentence-bucket support for aggregation.
Together with the within-MEG evidence attenuation diagnostic, Brennan therefore supports the source-attribution interpretation: contextual aggregation becomes measurable only when local stimulus-locked evidence is present.

\subsection{Training-time local--global coupling as an architectural contrast}
\label{app:coupling}

As an architectural contrast to frozen-logit score-space aggregation, we evaluate a training-time local--global coupling model under the same shortcut-controlled closed-set retrieval setting as the main experiments. This control tests whether a straightforward end-to-end context pathway can recover the contextual headroom exposed by the oracle diagnostic. The model has two streams: a local encoder produces window features for each fixed-length query, while a global stream summarises sentence-level neural context into a small set of fixed slots~\citep{locatello2006object,jaegle2021perceiver}. Local features attend to these global slots via gated cross-attention, and the resulting context update is added as a gated residual to the local stream~\citep{alayrac2022flamingo}. All variants use the same candidate pool construction, stimulus-identity splitting, preprocessing, and retrieval objective as the local-only baseline.

\textbf{Results.}
Table~\ref{tab:retrieval_zeroshot_combined_final} shows that this local--global coupling architecture does not yield reliable gains over the local-only baseline under the tested low-SNR retrieval setting.
The orthogonal variant degrades retrieval, whereas the non-orthogonal coupling remains close to the local-only baseline.

\begin{table}[!ht]
\centering
\small
\setlength{\tabcolsep}{6pt}
\caption{\textbf{End-to-end local--global coupling under shortcut-controlled retrieval on Gwilliams zero-shot.}
Retrieval rates are reported as percentages; MedR is reported in rank units.}
\renewcommand{\arraystretch}{1.15}
\label{tab:retrieval_zeroshot_combined_final}
\begin{tabular}{lcccc}
\toprule
\textbf{Model variant} & \textbf{R@1 (\%)} & \textbf{R@10 (\%)} & \textbf{MRR (\%)} & \textbf{MedR} \\
\midrule
Local-only baseline        & 44.0 & 73.0 & 54.0 & 2.0 \\
Local--global (orthogonal) & 25.0 & 58.0 & 39.0 & 6.0 \\
Local--global (non-orth.)  & 43.0 & 72.0 & 53.0 & 2.0 \\
\bottomrule
\end{tabular}
\end{table}

\textbf{Interpretation.}
A plausible explanation is an imbalance in stream optimisation: the local pathway provides stronger, more direct gradients for window-level retrieval, while the global stream is weaker or noisier early in training.
This can encourage the gating mechanism to downweight context injection, reducing the learning signal for the global stream.
Additionally, fixed-length windowing forces variable-length context into fixed-size slots, yielding noisy sentence features.
The orthogonality constraint further restricts how context interacts with local evidence, potentially disrupting the representation space required for retrieval under low SNR.
These hypotheses are consistent with the observed negative result.
The main role of this control is to contrast training-time context injection with the score-space GCB intervention, where contextual changes are observable without retraining or modifying the local encoder.

\subsection{Score-space visualisation of GCB}
\label{app:gcb_tsne}

\begin{figure}[!ht]
  \centering
  \includegraphics[width=0.7\linewidth]{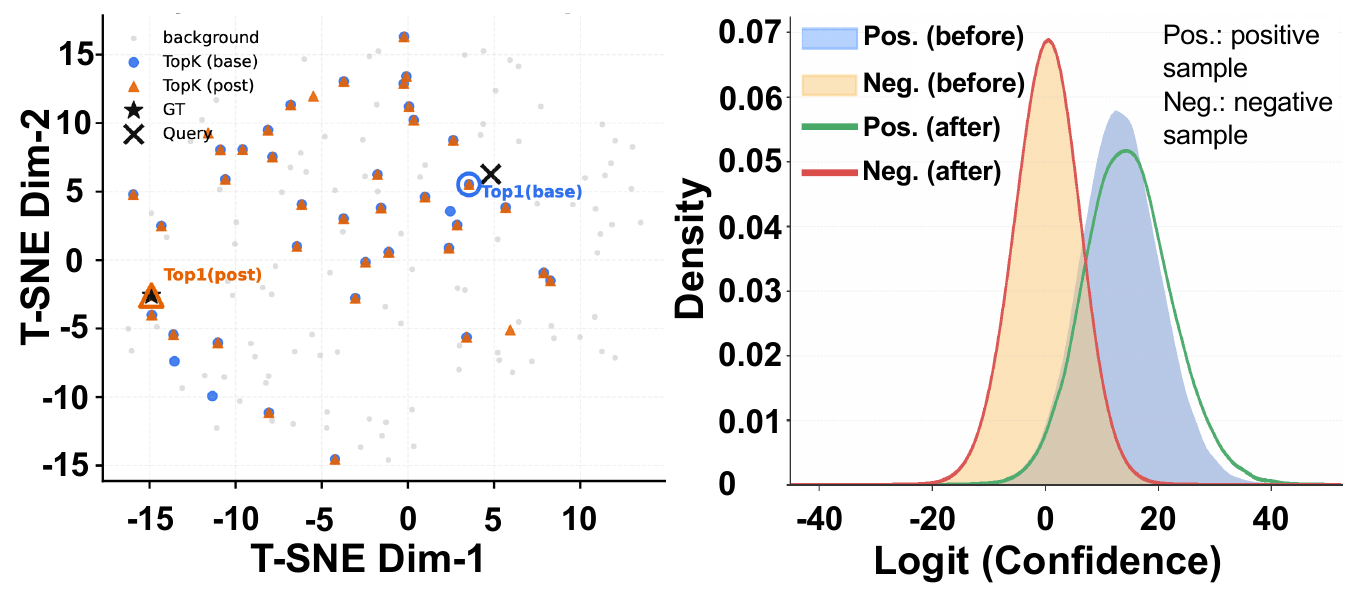}
  \caption{\textbf{Score-space visualisation of GCB while keeping embeddings fixed.}
  Left: A t-SNE visualisation of a query and its top candidates before and after GCB.
  The visualisation illustrates that GCB does not modify the embedding geometry; it only changes the ranking through a post-hoc logit correction.
  Right: Retrieval-logit distributions for positive and negative candidates before and after GCB.
  The positive candidate distribution shifts upward after GCB, while the negative distribution remains largely unchanged.}
  \label{fig:app_gcb_tsne}
\end{figure}

Fig.~\ref{fig:app_gcb_tsne} provides an illustrative score-space visualisation of GCB on a representative query.
The purpose is illustrative: the neural encoder and candidate embeddings remain fixed, while the retrieval decision can change after sentence-bucket evidence is aggregated in logit space.

\subsection{Qualitative spatial mixing visualisation}
\label{app:spatial_mix_viz}

\begin{figure}[!ht]
  \centering
  \includegraphics[width=0.5\linewidth]{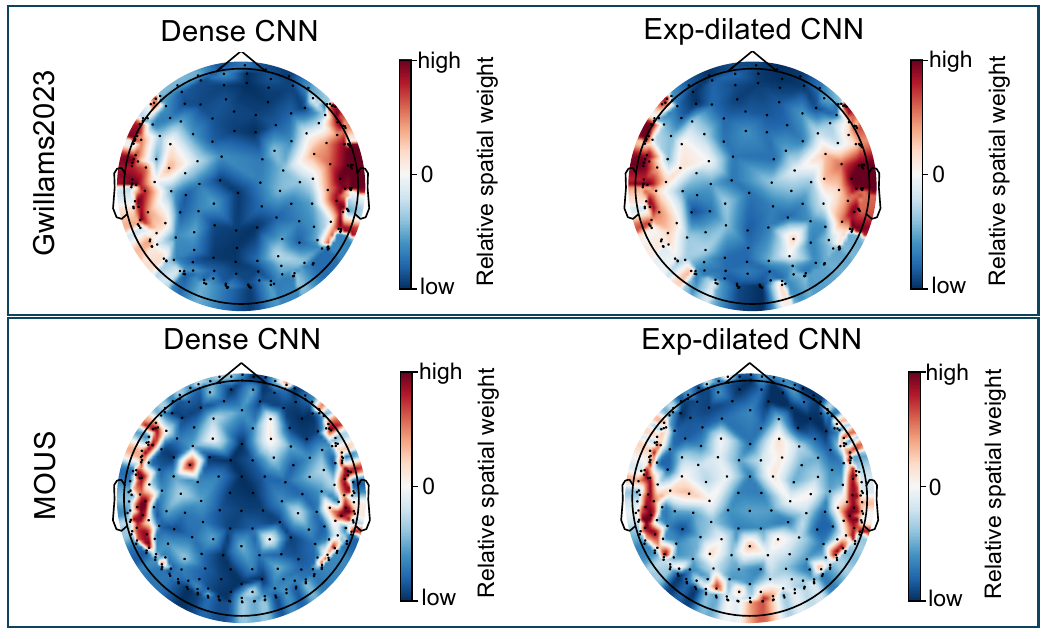}
  \caption{\textbf{Learned spatial mixing weights (qualitative).}
  Top: Gwilliams; bottom: MOUS. Left: Dense-TCNN; right: exp-dilated CNN.
  Colours indicate relative spatial mixing weight in the shared coordinate-conditioned frontend (red: high; blue: low).}
  \label{fig:spatial_weights}
\end{figure}

Fig.~\ref{fig:spatial_weights} visualises the learned sensor-space mixing weights in the shared coordinate-conditioned spatial frontend.
The topographies suggest a consistent geometric pattern, with relatively larger weights on bilateral lateral sensors across datasets and backbones.
These sensor-level weights should be read qualitatively: they may reflect sensor geometry, SNR differences, preprocessing choices, or regularisation, and are not direct neuroanatomical localisation estimates.
We report this figure for transparency and qualitative intuition.

\subsection{Token-level case study of GCB-induced retrieval changes}
\label{app:qual_examples}

We provide a token-level case study of how GCB modifies retrieval-decoded reconstructions on Gwilliams.
Each query's neural window is mapped to its top-ranked audio candidate before and after GCB, and the audio candidate's associated word is recorded.
This gives a token sequence per sentence that reflects which candidate windows the retrieval model selects.
The pre-GCB sequence (``Base'') uses the local retrieval logits; the post-GCB sequence (``Post'') uses the GCB-corrected logits.
This is a retrieval-side case study of which words change rank under GCB, and is not a generative reconstruction.

Table~\ref{tab:viz} shows representative sentences in two regimes: those where GCB improves token-level exact-match accuracy against the target, and those where it degrades accuracy.
The improvement cases illustrate that GCB tends to repair multi-word spans that the local model partially matched, consistent with cross-window evidence pooling promoting candidates from the correct sentence bucket.
The worsened cases show occasional regressions when sentence-bucket support is misaligned.
Token-level alignment between candidate-window content and the target sentence is approximate; we treat this analysis as qualitative.

\begin{table*}[ht]
\centering
\caption{\textbf{Token-level visualisation of retrieval-decoded reconstructions before and after GCB correction.}
\fixgood{Grey italic}: corrected tokens (Base wrong \(\rightarrow\) Post correct);
\fixbad{red underline}: tokens that became incorrect;
\stillwrong{blue underline}: still incorrect but changed.
\emph{Base, Post, and \(\Delta\) report token-level exact-match accuracy against the target sentence, computed at aligned word positions, and its change \((\Delta=\mathrm{Post}-\mathrm{Base})\).
Values are reported as proportions.}}
\label{tab:viz}
\scriptsize
\setlength{\tabcolsep}{4pt}
\renewcommand{\arraystretch}{1.15}
\begin{tabular}{p{0.26\textwidth} p{0.26\textwidth} p{0.26\textwidth} ccc}
\toprule
\textbf{Target} & \textbf{Base Output} & \textbf{Post Output} & \textbf{Base} & \textbf{Post} & \(\boldsymbol{\Delta}\) \\
\midrule
\multicolumn{6}{l}{\textbf{Improved}} \\
\midrule
Acres stood beside it and somehow using him as a yardstick for the scale of the thing made it easier to see &
but right wound car now on Perhaps folders and his the You the who up reached who that it his to see &
Roy right wound car now on Perhaps \fixgood{him as a yardstick for the scale of} reached who made it easier to see &
0.18 & 0.59 & +0.41 \\

He took another careful look over his shoulder and bent to pluck a yucca spear as if that were what drew his interest &
frame we rental characters look over his shoulder to shoulder to psi a a marked as if that were what originality being interest &
He took another careful look over his shoulder and \fixgood{bent to pluck a yucca spear} as if that were what \fixgood{drew} else interest &
0.52 & 0.87 & +0.35 \\

For some reason I thought of a giant clown face from a carnival ride but it scared the hell out me &
At some reason I She of a giant before head ridiculous and phone Charles toes it scared the hell out me &
For some reason I \fixgood{thought of a giant clown face from a carnival ride} but it scared the hell out me &
0.57 & 0.90 & +0.33 \\
\midrule
\multicolumn{6}{l}{\textbf{Worsened}} \\
\midrule
The gold was heavy on my back and I was all the way to the car before I realized I had left the keys on Charles &
The gold was heavy on my back and I was all the way to the car before I realized I would left the keys on Charles &
Hey gold was heavy on my back \fixbad{Right} I was all the way to the car before I realized I would left the keys on \fixbad{of} &
0.96 & 0.85 & \(-0.12\) \\

They never bothered trying to explain how Acres made it into the country without a ticket on a plane or a boat &
They never bothered younger to explain how was made it into the country without a ticket stick a actually or a do &
They if bothered younger to explain how was made it into the \fixbad{Roy without a parts when} a actually or a do &
0.77 & 0.64 & \(-0.14\) \\

The gold was heavy on my back and I was all the way to the car before I realized I had left the keys on Charles &
They he was heavy up my them all of its of the pines to out car before I realized I had left the keys me bound &
They he was heavy up patrol them all of its of the pines to out car before I realized I \stillwrong{proposition left was was} me bound &
0.54 & 0.38 & \(-0.15\) \\
\bottomrule
\end{tabular}
\end{table*}

\end{document}